\pdfoutput=1
\documentclass[11pt]{article}

\usepackage[preprint]{acl}
\usepackage{enumitem}
\usepackage{xcolor}
\usepackage{times}
\usepackage{latexsym}
\usepackage{subcaption}
\usepackage{graphicx}
\usepackage{caption}
\usepackage{listings}
\usepackage{float}
\usepackage{amsmath}

\usepackage[T1]{fontenc}
\usepackage[utf8]{inputenc}

\usepackage{microtype}

\usepackage{inconsolata}

\definecolor{dark green}{RGB}{0,100,0}
\title{Mining Contextualized Visual Associations \\ from Images for Creativity Understanding}
\author{
  Ananya Sahu, Amith Ananthram, Kathleen McKeown \\
  Columbia University \\
  \texttt{as5957@columbia.edu, amith@cs.columbia.edu, kathy@cs.columbia.edu}
}

\begin{document}
\maketitle
\begin{abstract}

Understanding another person’s creative output requires a shared language of association.  However, when training vision-language models such as CLIP, we rely on web-scraped datasets containing short, predominantly literal, alt-text.  In this work, we introduce a method for mining contextualized associations for salient visual elements in an image that can scale to any unlabeled dataset.  Given an image, we can use these mined associations to generate high quality creative captions at increasing degrees of abstraction.  With our method, we produce a new dataset of visual associations and $1.7$m creative captions for the images in MSCOCO.  Human evaluation confirms that these captions remain visually grounded while exhibiting recognizably increasing abstraction.  Moreover, fine-tuning a visual encoder on this dataset yields meaningful improvements in zero-shot image-text retrieval in two creative domains: poetry and metaphor visualization.  We release our dataset, our generation code and our models for use by the broader community.

\end{abstract}

\section{Introduction}

% \aanote{Can we add a first paragraph that leads with creative image understanding -- perhaps referencing some of the work from the cognitive sciences that you've found that explains the role that association plays?  Then in paragraph two we can contrast this against the dominant paradigm in VLMs: CLIP.  And then paragraph three should be our corpus generation (contrasting against related work), paragraph four our evaluation setup, paragraph five our conclusions.} %done

%para 1

% \aanote{Potential alternative to paragraph 1}

\begin{figure}[ht]
    \centering
    \includegraphics[width=\columnwidth]{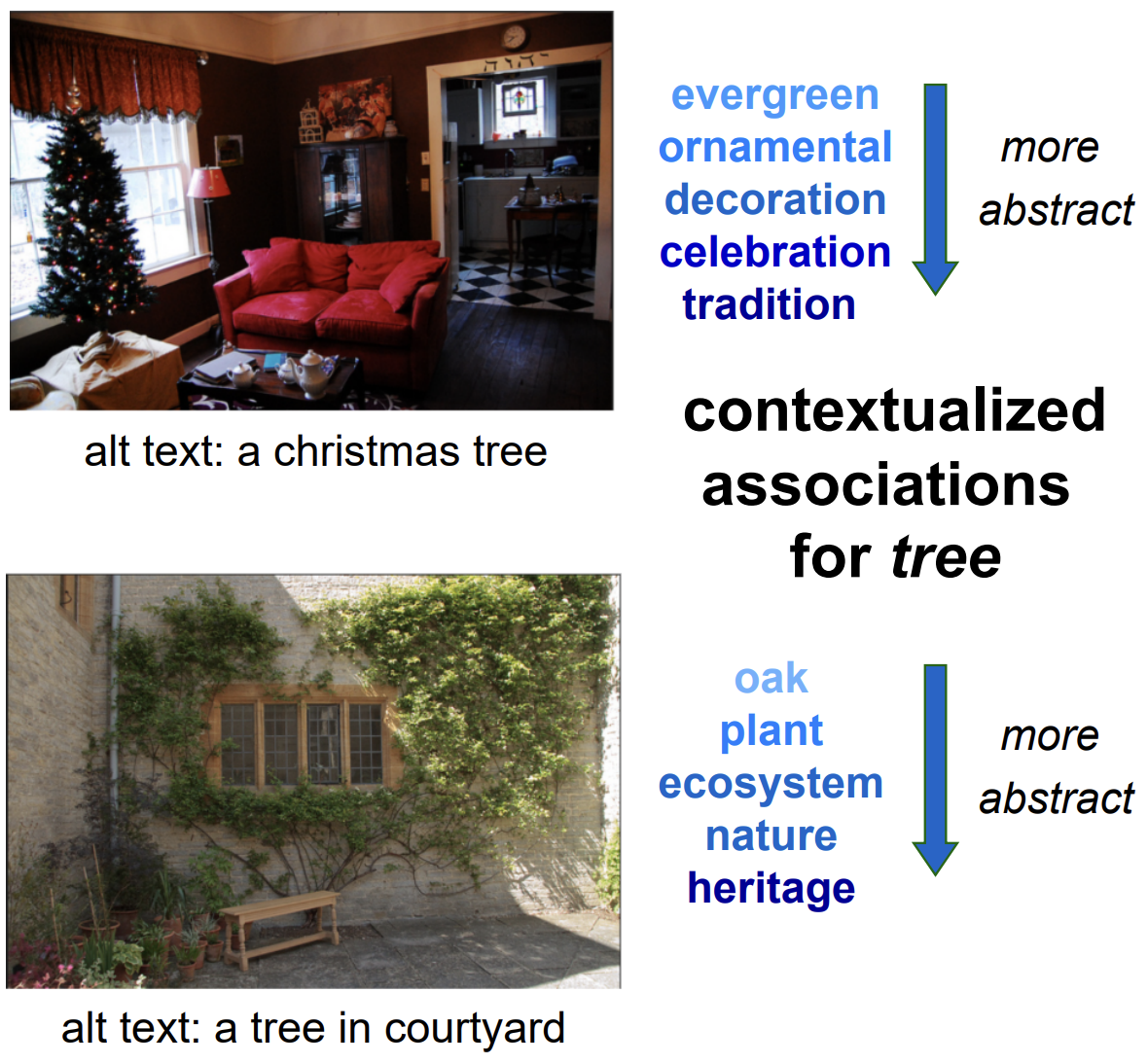}
    \caption{Two images depicting \textit{trees} in different settings. Their alt-text makes no mention of the diverse concepts that each tree evokes.  Using our method, we are able to mine contextualized associations at degrees of abstraction that extend beyond literal description.
    % On the right we show contextual associations for \textit{tree} showing increasing abstraction from concrete to abstract concepts based on the its visual contexts in each image which are useful for interpreting the images beyond their literal descriptions
    } 
    \label{fig:teaser}
\end{figure}

We make sense of visual art through shared associations \cite{gombrich2023art}.  Studies from the cognitive sciences have shown that these associations come from our collective biological, social, cultural and environmental contexts \cite{ward2010cognition}.  For example, skulls evoke death in many Western viewers.  Consequently, creative vision-language tasks like art interpretation or image-to-poetry generation require models that can leverage these same associations \cite{huang2016visual,hu2020makes,liu2018beyond,lu2022artcap}.  

% Creative image understanding involves the ability to recognize and understand ideas conveyed in an image. Creative vision language tasks such as visual storytelling, generating captions of artworks, and image-to-poetry generation involve an added complexity to models requiring them to reason about the images and exhibit creativity in the output text generations \cite{huang2016visual,hu2020makes,liu2018beyond,lu2022artcap}. Studies informed from cognitive science suggest the outcome of creativity comes from social, cultural, situational and individual contexts \cite{ward2010cognition}. Therefore creative image understanding tasks require models to have additional world and commonsense knowledge.\cite{chakrabarty2023creative}.        

%para 2

However, while training on image-text pairs scraped from the web has yielded powerful models 
% \kmnote{More formal in a paper. Change "like" to "such as"}
such as CLIP that are able to adapt to many downstream tasks, %ranging from image captioning to visual question answering, 
research has found that they often fail to achieve similar zero-shot performance in tasks where the domain is largely different from their pre-training data \cite{menon2024task}. This is especially true in creative domains.  In poetry and metaphor visualization, CLIP's capabilities are limited \cite{guljajeva2023explaining}.  We hypothesize that this is because the text seen during its pre-training is predominantly short alt-text which does not explicitly include any associations for its accompanying imagery (see Figure \ref{fig:teaser}). 
% \kmnote{This sentence somewhat convoluted. "make" and "explicit" separated by a long distance. How about: "which do not explicitly include any associations for its accompanying imagery" (I don't think you would expect it to have your assocations, so I would definitely drop "our" }

% .  However, studies have shown that CLIP Particularly in domains requiring creativity, like poetry or visual metaphors, zero-shot CLIP capabilities are limited . 

%para 3 our corpus generation contrasted against related work  Can we replace this with a more detailed description of our corpus generation technique and in so doing motivate the usefulness of having associations/descriptions for the same image at various levels of abstraction.  Perhaps you can find something to cite here?
Prior work has improved vision-language models (VLMs) by training on synthetic captions with fine-grained detail, resulting in more nuanced image understanding %enabling more nuanced vision language-alignment 
\cite{chen2024sharegpt4v,fanimproving,lai2024veclip}.  This has produced meaningful performance gains in classification and cross-modal retrieval tasks in non-creative domains. 

In our work, we extend this effort to creative domains.  We develop a method for mining contextualized visual associations for the salient elements in 
% \kmnote{I find "any" too strong here. How about just "an"}
an unlabeled image.  Here, we define contextualized associations as concepts related to a particular visual element based on broader scene context (e.g., ``celebration" for the Christmas tree in Figure \ref{fig:teaser}).  
% \kmnote{Note that you have a lot of typos. I have been correcting as I read, but you should do spell check and careful read. below "produced" should be "produce". Also this is the point you introduce synthetically produced creative captions. This synthetic generation should come before the rewriting sentence in the abstract. }
Then, we use these mined visual associations to synthetically produce creative captions for each image at increasing degrees of abstraction, informed by Hayakawa's ``ladder of abstraction" from linguistics \cite{hayakawa1967language}.  This results in captions that remain grounded to an image while making explicit the associations that the image evokes.  

% hey ananya, i'll tackle kathy's suggestions in the abstract / intro, can you see my slack messages?  and then take a stab at fixing the related work?

% In our work, we address challenges in aligning CLIP, for image and creative text pairs. We hypothesize adding contextualized visual associations will improve performance in downstream creative tasks. We first obtain contextual conceptual associations from a Large Language Model for all salient objects within an image and caption pair at varying abstraction levels. We then synthetically generate creative captions by prompting a Vision Language Model to rewrite the original caption using our obtained conceptual associations. This allows for the caption to be grounded in the image as opposed to free-form prompting of generating creative captions for an image which often contain hallucinations. 

%para 4 evaluation set up 
% \kmnote{since you are talking about scaling here, you don't need to include words like "any" earlier}
% AA: addressed by removing any above
Our data generation process is general purpose and can be arbitrarily scaled to any unlabeled corpus of images. We validate the quality of the resulting creative captions through 1) human evaluation and 2) testing the ability of a visual encoder fine-tuned on our synthetic dataset to adapt to two creative vision-language tasks: image-to-poetry retrieval and linguistic metaphor-to-visual metaphor retrieval \cite{liu2018beyond,chakrabarty2023creative}.  We find that our synthetic captions reflect increasingly creative abstraction that aligns well with human judgment without introducing hallucination.  Moreover, fine-tuning on these captions improves zero-shot multi-modal retrieval in both of our creative vision-language tasks.

In summary, the contributions of our work are:

\begin{itemize}
            \item A novel approach for mining contextualized associations for visual elements in unlabeled images at increasing degrees of abstraction
            %; given an image, we identify salient visual elements and then prompt a text-only frontier model to generate associations for each visual element based on a detailed description of its surroundings
            % from salient objects in image caption pairs: Use a VLM (Molmo) to generate a detailed caption of a given image which we provide as context to the LLM, gpt4o-mini and prompt to generate associations based in the input context,  gradually abstracted from the original salient word. To obtain salient objects, we extract the most meaningful words of the original caption (nouns, verbs, adjectives) and keep the five most concrete words based on word concreteness ratings \cite{brysbaert2014concreteness}. We take inspiration from the linguistic framework of the ladder of abstraction \cite{hayakawa1967language} in which language is thought to be categorized as different rungs on the ladder, with the bottom most being concrete and the top most being abstract. The generated visual associations at each abstract distance then get input to a VLM to rewrite the original caption, replacing the original salient word with our mined contextual association to generate our creative captions.  \kmnote{Yes should be 2-3 lines max} 
             % \aanote{Need to make this shorter; don't think we need to specify the models we use, etc; just the high level of the method and what it produces}
            % \aanote{Can you we describe the process for mining your visual associations as it's own contribution here?  And then talk about the dataset as point #2 below}
            \item 
            % \kmnote{I would stress here that it's a dataset. Perhaps " A new dataset, extending MSCOCO with mined visual..."}
            A new dataset, extending MSCOCO with increasingly abstract visual associations and accompanying high quality creative captions 
            \item 
            A human evaluation of our dataset, validating both the increasing abstraction and visual grounding of our synthetic captions
            \item 
            An evaluation of CLIP, fine-tuned on our dataset, showing improved performance for multiple creative cross-modal retrieval tasks
            % \item Evaluate CLIP for aligning these creative caption pairs and images, and train a modified CLIP with creative image and text pairs at multiple abstraction levels using our corpus and distance labels for task guidance 
            % \item Show improved performance for zero shot image creative text cross modal retrieval for multiple corpora with our dataset and model \kmnote{These last two bullet points are not separate. Also all your bullets should be parallel structure. bullet 2 is a noun phrase. Bullets 3 and 4 should also be a noun phrase. How about something like "An evaluation of CLIP, fine-tuned on our dataset, showing improved performance for multiple creative cross modal retrieval tasks. "}
        \end{itemize}

Additionally, we release our dataset, generation code and models for use by the broader community: \url{https://github.com/ananya-sahu/mining_visual_associations}.

\section{Related Work}

% \aanote{I don't think we need all this background on VLMs.  Let's just focus on the specific VLM we use, CLIP.  Describe what we know about its pretraining data and the charactersitics of it.  And then maybe a sentence about why its important (CLIP is a backbone for many VLMs).  So compress paragraphs one and two here into a single paragraph that hits all of those notes.  Paragraph three is good as is.  Remember we only want to talk about other work insofar as it relates to our research focus here.} 

% Vision Language Models are multimodal modals that can intake images and text and undergo vision and natural language tasks. They usually consist of a vision encoder, modality connector component, and a large language model. 

% \aanote{I am not sure we need to cover CLIP and its biases here in the Related Work.  IMO what's here is the motivation for our work as discussed in the Introduction.  Here, let's cover synthetic data generation for vision-language modeling (the third paragraph).  Also let's move the section about conceptual associations and creativity to the first part of the Related Work}

\subsection{Conceptual Associations and Creativity}

% \aanote{This stuff is great -- perhaps draw from it for your introduction paragraph.  We want to say "cogsci says we do creativity like this but the way we train models is X, we make Y change to get us closer to the way humans work"}

Cognitive science has shown that creativity involves associative thinking \cite{ward2010cognition}.  Often, this entails linking together related concepts through abstraction \cite{beaty2023associative}.  In NLP, attempts to understand poetic language, including metaphor, simile and emotion, have required external associative knowledge to help make sense of implicit meaning \cite{chakrabarty2022flute}.  A common method for incorporating such knowledge is through the use of association lexicons.  Previous studies have collected rich lexicons for the colors and emotions evoked by different words through painstaking human annotation \cite{mohammad2013colourful,mohammad2013crowdsourcing}.  These were complemented by efforts at automating association mining through word embeddings \cite{bolukbasi2016man,hu2019natural}.  
% Recently, \citet{abramski2024llm} used generated free associations from LLMs to reveal the structure of their conceptual knowledge.  
In contrast with this prior work, we present a method for automatically mining \textit{contextualized} associations, where the same word's related concepts vary based on its surroundings.  Moreover, while previous lexicons have typically focused on text, our associations are \textit{visually contextualized}, extending association mining to a new modality.
\subsection{Synthetic Image-Text Data}

Due to the strength of current VLMs, recent work has exploited synthetic data to improve the downstream performance of image encoders on vision-language tasks \cite{zheng2024dreamlip,kong2024hyperbolic,xiao2024flair,liu2024clips,yang2023alip}. Studies have found that generated captions can be longer and more descriptive of images than their naturally occurring references \cite{chen2024sharegpt4v,sharifzadeh2024synth}. Some have even shown that training on such captions can yield higher performance than training on those from human annotators  \cite{santurkar2022caption}.
% Recently \citet{liu2024clips} introduced a framework to generate synthetic captions conditioned on existing image text pairs. Another work, \citet{yang2023alip}, incorporates supervision from both synthetic and raw caption data and has shown enhanced performance in downstream tasks of zero-shot image text retrieval.  
While exciting, the focus of much of this work has been on improving the performance of VLMs on standard image understanding tasks.  In our work, we expand this line of inquiry to include creative domains.  Building on our method for mining contextualized associations, we generate a corpus of creative captions and show that training on these captions yields significant improvements on zero-shot image-poetry and image-metaphor retrieval.  

% In contrast, in our work we utilize visual contextual associations to generate synthetic creative captions to improve downstream creative image text understanding. Using our dataset, we show improved zero-shot downstream performance on creative image text retrieval tasks in unseen domains of image-poetry pair and visual metaphors. 
      
% \aanote{Can you add a sentence here that is something like, "In contrast, we do X".  We want one for the prior paragraph too.} (done)
 % \aanote{I think this last paragraph can be cut -- at least the CLIP task bias stuff.  Perhaps just take the last sentence and fold it into your prior paragraph at the end where you contrast against the prior work.} (done)
%  In our work we leverage the usage of conceptual associations for creative understanding and synthetic data generation to improve image text alignment in CLIP. 
% % \aanote{The prefix stuff is more of an implementation detail -- can definitely talk about task bias but don't need to address prefix tuning it here.}
% Additionally, inspired by \cite{menon2024task} we explore CLIP's biases in understanding abstractness for image creative text pairs. Using our dataset, we show improved zero-shot downstream performance on creative image text retrieval tasks in unseen domains of image-poetry pair and  visual metaphors. 

% and emotion understanding in artworks. commenting this out for now

\section{Generating Abstracted Captions}
\begin{figure*}[h]
    \centering
    \begin{subfigure}{\linewidth}
        \centering
        \includegraphics[width=1\linewidth]{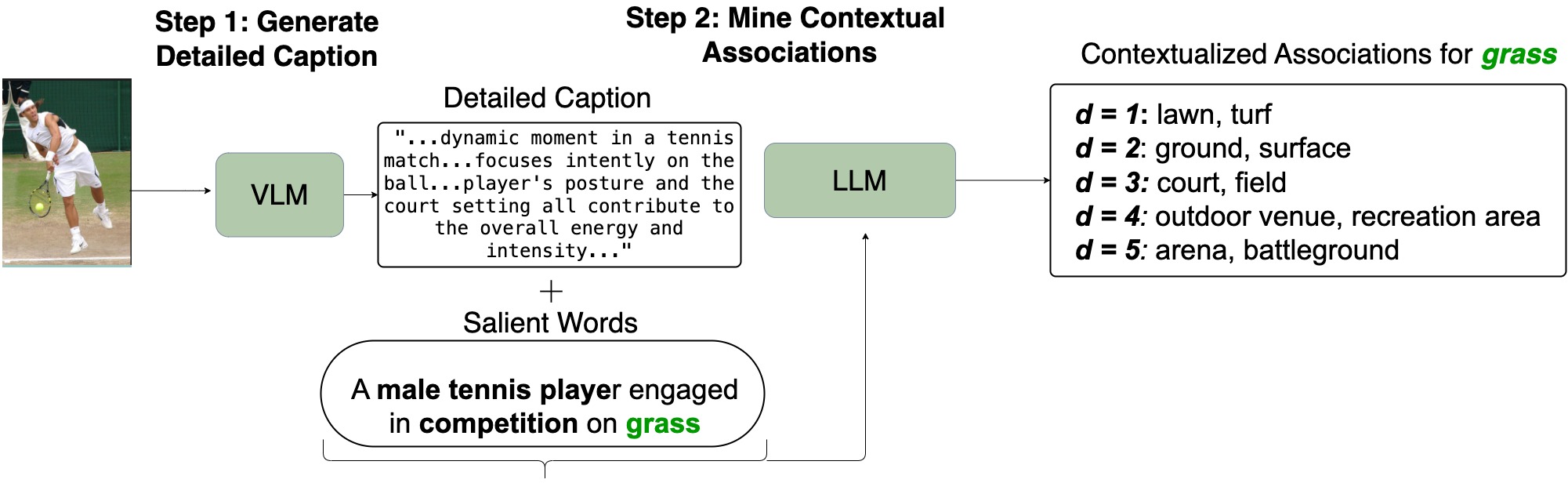}
        \caption{Mining contextualized associations for \textit{grass}.}
    \end{subfigure}
    
    \vspace{0.5cm} % Adjust spacing between images
    
    \begin{subfigure}{\linewidth}
        \centering
        \includegraphics[width=1\linewidth]{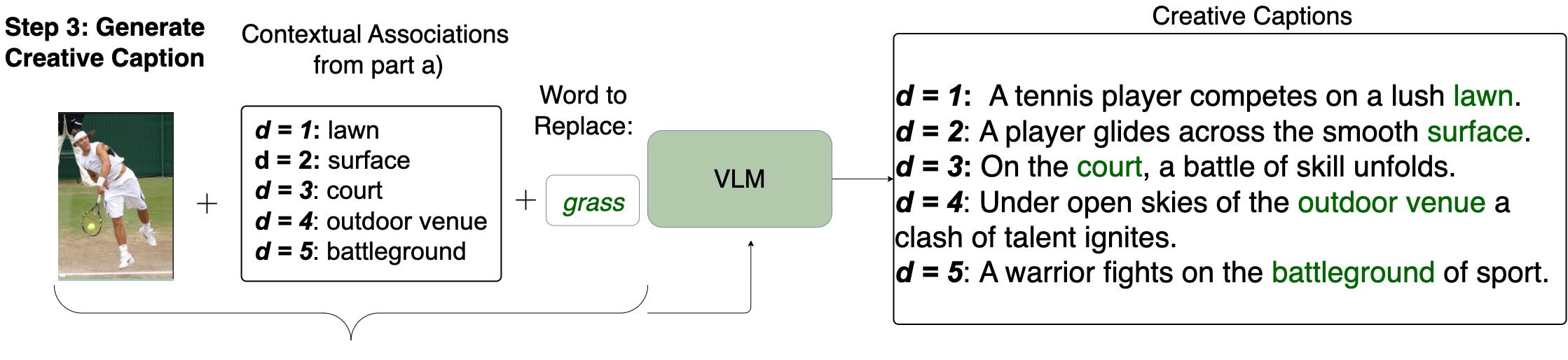}
        \caption{Generating creative captions using contextualized associations for \textit{grass}.}
    \end{subfigure}
    
    \caption{Our method for mining contextualized associations and generating creative captions with increasing abstraction.  In \textbf{Step 1}, given an image, we prompt a VLM to generate a detailed caption.  Then, in \textbf{Step 2}, we prompt an LLM to mine associations for each of its salient visual elements at increasing degrees of abstraction.  Finally, in \textbf{Step 3}, we prompt a VLM to generate synthetic creative captions using our mined associations.} 
    % \aanote{Can you make the text in this figure bigger?  It shoudl be at least as big as the text in the paper.  And once you've done that, can you move it to the same page as the methods and reference the different pieces of it in the methods?}%done needed to remove the second image though to fit in 1/3 of page and have font bigger  
    
    % \aanote{Some notes: 0) Can you number different stpes in a) and b) so we can refer to them from the text? 1) can you make all of the text bigger? You can reclaim some space by replacing "Distance" with "D1", "D2", etc.  Also instead of "dense" lets say "Detailed Captions" as dense captions mean something else in the literature (I just discovered this).  One way to perhaps reclaim some space so you can fit more text in is to include the images just once and then use them for both parts A and B. 2) Also, in the final formatting, let's make sure this image is on the same page as the methods section.}
    %I tried to make the text as big as possible for captions, and associations but lmk if it needs to be bigger still
   
    \label{fig:generation diagrams}
\end{figure*}

% \subsection{Dataset Creation}

Given an image $I$ featuring visual elements $V_I$, we mine a set of contextualized associations $A_d(v_j)$ for each $v_j \in V_I$ at increasing degrees of abstraction, $d \in D$.  Then, using these associations, we generate a set of captions $C_{d}(I)$ for each image $I$ that reflect the specified degree of abstraction, $d$.

In this work, we define contextualized associations as concepts that are related to the specified visual element $v_i$ based on its broader scene context.  For example, in Figure \ref{fig:teaser}, a \textit{tree} outside evokes different associations than an indoor Christmas tree in many Western viewers -- these associations are mediated by each tree's surroundings.

% To create our creative captions dataset we obtain contextual associations for the most salient aspects of the image and rewrite the caption with the associations. We define contextual associations as words that pertain to the salient object in the image but also take into consideration the other objects present within the image. For example, as depicted in Figure 1, the word tree in an outdoor nature setting will have different associations than an indoor Christmas tree due to their differing contexts, where context is defined by the setting and other objects present in the image.

%We obtain the contextual associations for salient aspects of the image at multiple distances and rewrite original captions into creative captions. The distances \( \alpha \)  are specified as: 

% We define $5$ distances of abstraction $d$, inspired by the linguistic concept of ladder of abstraction in which language is thought to be categorized in as different rungs on ladder, with the bottom most being concrete and the top most being abstract \cite{hayakawa1967language}:
We define five degrees of abstraction $d$, inspired by Hayakawa's ``ladder of abstraction" from linguistics \cite{hayakawa1967language}:
% \aanote{include citation for where these are from}:

\begin{enumerate}
    \item \textbf{Near Synonyms} ($\mathbf{d = 1}$): Close in meaning or form (e.g., Ball → Sphere).
    \item \textbf{Slight Abstractions} ($\mathbf{d = 2})$: Slightly broader category (e.g., Ball → Toy).
    \item \textbf{Broader Context} ($\mathbf{d = 3})$: Indirect, but linked through situational and emotional context (e.g., Ball → Game).
    \item \textbf{Conceptual Associations } ($\mathbf{d = 4}$): More abstract or thematic (e.g., Ball → Competition). 
    \item \textbf{Full Abstractions} ($\mathbf{d = 5}$): Highly abstract or metaphorical (e.g., Ball → Journey).
\end{enumerate}    
\subsection{Mining Contextualized Associations}

Given an image \( I \) with a short caption \( c_{short} \), first, we generate a detailed caption \( c_{detailed} \) using an off-the-shelf vision-language model (VLM).  Then, we extract salient visual elements \( v_1, \dots, v_n \in V_I\) by identifying nouns, adjectives, and verbs in the short caption \( c_{short} \) with high concreteness ratings according to a lexicon \cite{brysbaert2014concreteness}.  

As large language models are trained on language that is much richer than the language typically found in image alt-text, they function as high quality repositories of common associations, especially when conditioned with complete scene context \cite{tsimpoukelli2021multimodal}.  Thus, we prompt a text-only frontier language model with both our detailed caption  \( c_{detailed} \) and our extracted visual elements $V_I$ to mine contextualized associations $A_d(v_j)$ for each $v_j \in V$ at every degree of abstraction $d$. We include the full prompt in \ref{sec:appendix_associations_prompt}.

\begin{figure*}[h!] % positioning: h = here, t = top, b = bottom, p = page of floats
    \centering
    \includegraphics[width=\linewidth]{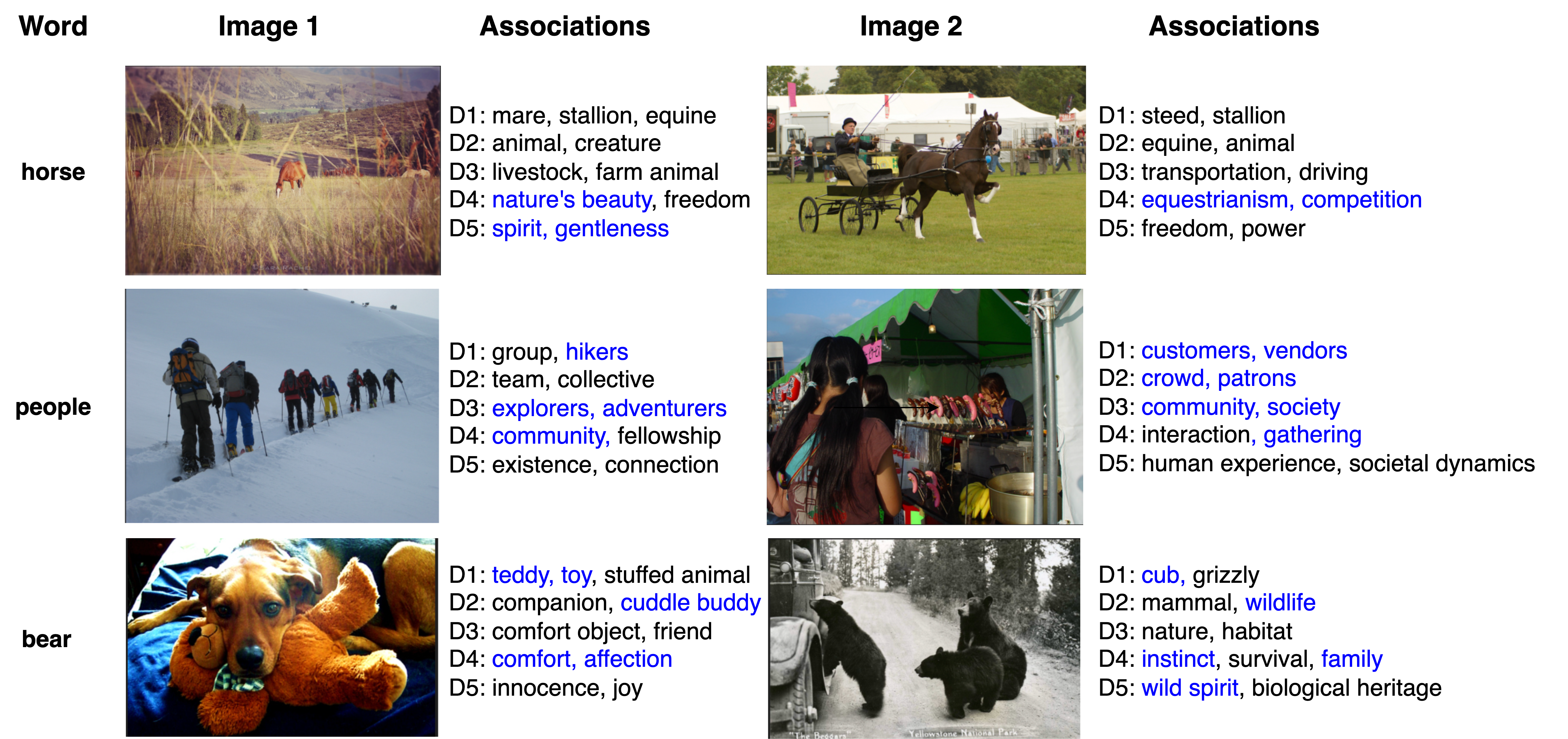} % adjust width as needed
    \caption{Examples from our corpus. For each word, we depict its contextualized associations at increasing degrees of abstraction for two representative images. Word associations change based on scene context.}
    \label{fig:datasetexamples}
\end{figure*}

Given an image \( I \), a salient visual element \( v_j\) and its conceptual associations \( A_d(v_j) \) at degree of abstraction $d$, we prompt a VLM to generate a creative caption $c_{creative}$ for each association in \( A_d(v_j) \). We include the full prompt in \ref{sec:appendix_abstracted_caption}.
% We also include in the prompt the same abstraction distance levels definitions used for obtaining the associations. 

% \aanote{This is fine here but let's try to do it with some mathematical notation and in particular include the objective we're optimizing.  Doesn't need a lot of space but should be precise}

\section{Experiments}

% \aanote{Add a section here with the experimental specifics that are in your methods right now.}

\subsection{Corpus Generation}

For our corpus of images with short captions ($I, c_{short}$), we use Microsoft Common Objects in Context (MSCOCO) \cite{COCO} due to its extensive study in vision language modeling.  We note, however, that our method can be applied to any corpus of unlabeled images for which we can obtain high quality short captions; given the strength of current VLMs, this includes most image corpora  \cite{bordes2024introduction}.  To extract salient visual elements from each short caption $c_{short}$, we employ \texttt{SpaCy}'s part of speech tagger and filter words based on their concreteness ratings using the lexicon from \citet{brysbaert2014concreteness} (requiring a minimum concreteness of $3$). We produce detailed descriptions of each image using \texttt{Molmo-7B-D-0924} \cite{deitke2024molmo}. We use text-only \texttt{GPT-4o-mini}\footnote{specifically \texttt{gpt-4o-mini-2024-07-18} last updated January 2025} 
% \aanote{Can you include the date range where you generated these assocations?} added in the model with date version 
to mine contextualized associations at different degrees of abstraction for each image's salient visual elements based on its detailed description.  
Finally, we use \texttt{Molmo-7B-D-0924} once again to generate a creative caption $c_{creative}$ for each extracted visual association. Example creative captions are shown in Figure \ref{fig:abstract_caption_examples}. In total, we produce $1,671,835$ creative captions for $\text{MSCOCO}_{train}$ and $102,552$ creative captions $\text{MSCOCO}_{validation}$ respectively. 

% \subsection{Evaluating CLIP on Creative Captions Dataset}
% To examine issues CLIP faces in associating creative image captions with their source images, we examine CLIP in two settings, 1) measuring CLIP similarity between original captions and their images and comparing them with measuring CLIP similarity between creative captions from our dataset and their corresponding images and 2) measuring CLIP similarity between hallucination captions and their images from the FOIL dataset (cite) and comparing them with measuring CLIP similarity between creative captions from our dataset and their corresponding images.

\subsection{Human Evaluation}

In order to validate our method for mining contextualized visual associations and generating creative captions, we conduct a human evaluation of our synthetic dataset.  We recruit five native English speakers to annotate a random sample of our corpus, answering two questions of interest:  First, how visually grounded (i.e. free of mistakes / errors / hallucinations) are the creative captions?  And second, how well do the generated creative captions reflect increasing abstraction?

To evaluate visual grounding, for $100$ creative captions, % with their respective images, 
we ask annotators to label whether the caption is completely contradictory to or not relevant to its image (rating of $1$), contains many erroneous details but still describes its image (rating of $2$), is an almost perfect caption with minor errors (rating of $3$) or represents a perfect caption where there are no errors (rating of $4$).  
% The goal of this task is to understand whether more abstract captions prone to hallucination-like errors.  

To evaluate abstraction, for each of $100$ images, we ask annotators to rank six of its captions in order of increasing abstraction: its original caption and one creative caption from each of our five abstraction degrees, presented in randomized order.  

We include our task instructions and screenshots of our annotation interfaces in section \ref{sec:appendix_interface}.

% each of the five distance level creative captions associated with the image as well as the original caption  abstractness where the first caption in the ranking order would contain the most literal description of the image (ranking of 1) and the last caption in order would contain the most abstract description of the image (ranking of 6). The goal of this task is to a) assess if our creative captions are indeed more abstract than the original captions as well as b) if our creative captions reflect the abstractness levels given.  

\begin{figure*}[ht]
    \centering
    \includegraphics[width=1.8\columnwidth]{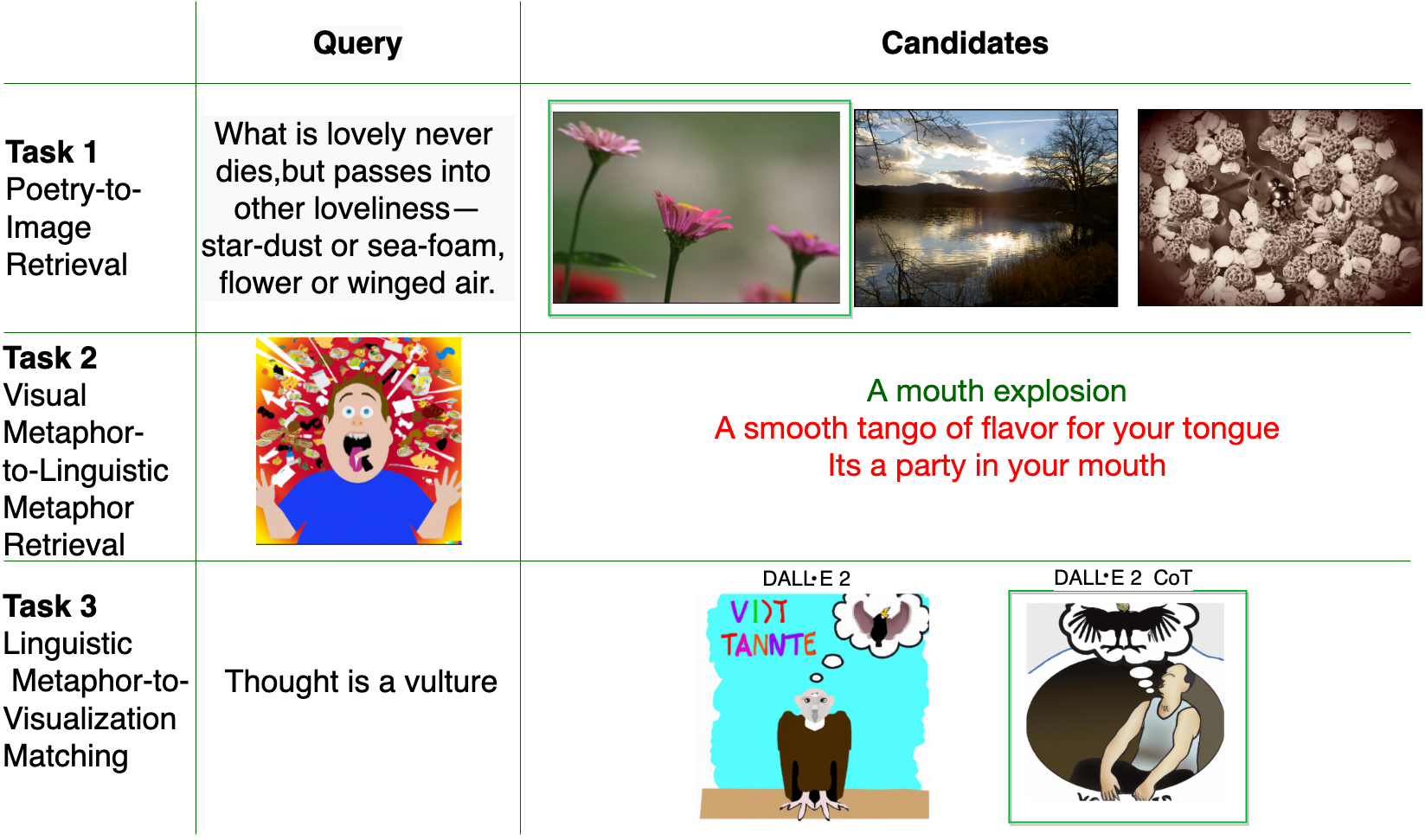}
    \caption{Examples from each of our three evaluation tasks.  Correct answers are highlighted in green.
    } 
    \label{fig:evaluation task examples}
\end{figure*}

\subsection{Automatic Evaluation}

In addition to a human evaluation, we validate our method for mining contextualized associations and generating creative captions by fine-tuning a pre-trained visual encoder on our corpus of creative captions for MSCOCO.  In particular, we expand \texttt{OpenCLIP-ViT-B/32}\footnote{pre-trained on the \texttt{laion2b\_s34b\_b79k} dataset} with a learnable prefix specific to each of our five degrees of abstraction $d \in D$ \cite{li2021prefix,menon2024task}.  Keeping the rest of the model frozen, we update only these prefix embeddings by optimizing CLIP's contrastive image-text matching loss on our corpus.  We fine-tune these weights for a single epoch.

% \aanote{I think you need a little more detail about each of these evaluation tasks.  What is the input?  What is the output?  How is the task structured?  For example, in HAIVMet you have to retrieve the correct image right?  Let's make sure each one is explained clearly (and concisely).  We might also add a figure that shows an example from each task.}

We compare our baseline, \texttt{OpenCLIP-ViT-B/32} without any fine-tuning, to our fine-tuned model at all five different degrees of abstraction -- that is, using each of our five learned abstraction prefixes.  We evaluate image-text similarity scores from these models on three zero-shot tasks constructed from datasets in two creative domains: 
% When evaluating our fine-tuned model, we consider 
% the resulting fine-tuned model against baselines of zero-shot CLIP. For our models we conduct all evaluation tasks in all abstract distance settings of D = 1,2,3,4,5. 
% and CLIP fine-tuned on captions directly prompted using MOLMO without any associations inputted.
% The two cross model retrieval tasks we consider are:
% We evaluate on the task of cross modal retrieval (image to text and text to image) in the following datasets: 
\begin{itemize}
\item  \textbf{Multi-Modal Poem (MultiM-Poem)} \cite{liu2018beyond}: Contains $8,292$ images from Flickr paired by English majors with short poems (around $7$ lines) from several online poetry sites\footnote{Foundation3, PoetrySoup4, best-poem.net and poets.org}.  We use MultiM-Poem for \textbf{Task 1}, poetry-to-image retrieval: given a poem, retrieve its corresponding image. 

% \aanote{from where?}. % added origin of images and poems 

% We use the MultiM-Poem dataset for task 1 of our evaluation
\item  \textbf{HAIVMet} \cite{chakrabarty2023spy}: Contains $1,540$ linguistic metaphors paired with both incorrect, overly literal, visualizations generated by DALL·E2 and correct, appropriately metaphorical, visualizations generated by DALL·E2 through chain-of-thought.  We use HAIVMet for \textbf{Task 2}, visual metaphor-to-linguistic metaphor retrieval, and \textbf{Task 3}, linguistic metaphor-to-visualization matching.  
\end{itemize}

For our retrieval tasks, we report recall at $k=1, 5, 10, 20$ as well as the average rank of the correct text or image among all candidate texts or images (where lower is better).  For our matching task, we report how often the correct visualization is chosen over the incorrect visualization. We provide examples of each evaluation task in Figure \ref{fig:evaluation task examples}.

\section{Results and Discussion}
% \subsection{Evaluating CLIP on Creative Captions Dataset}
% To examine issues CLIP faces in associating creative image captions with their source images, we examine CLIP in two settings, 1) measuring CLIP similarity between original captions and their images and comparing them with measuring CLIP similarity between creative captions from our dataset and their corresponding images and 2) measuring CLIP similarity between hallucination captions and their images from the FOIL dataset (cite) and comparing them with measuring CLIP similarity between creative captions from our dataset and their corresponding images.
\begin{table}[t] % "t" keeps it at top of a column, use [h] for here
\centering
% \small % makes it fit better
\setlength{\tabcolsep}{6pt} % reduce column spacing
\begin{tabular}{|l|c|c|c|c|c|}
\hline
\textbf{Split} & D1 & D2 & D3 & D4 & D5 \\ 
\hline
Train & 59.1 & 69.3 & 77.9 & 79.6 & 80.0 \\
\hline
Val   & 62.2 & 71.9 & 79.8 & 81.8 & 81.9 \\
\hline
\end{tabular}
\caption{For each salient word and its images, the percentage of mined associations that are unique to an image, averaged over all the salient words in our corpus and calculated at each degree of abstraction.  Most associations are unique to a particular image's visual context and become increasingly so with abstraction.}
\label{tab:unique_associations}
\end{table}

\begin{table}[h]
    \centering
    \begin{tabular}{|c|c|}
        \hline
        \textbf{Abstraction} & \textbf{\% 
 with Grounding  $\mathbf{\geq 3}$}\\ 
        \hline
        Captions at $d=1$ & $90.1$\\
        \hline
        Captions at $d=2$ & $86.8$ \\
        \hline
        Captions at $d=3$ & $93.2$ \\
        \hline
        Captions at $d=4$ & $76.9$ \\
        \hline
        Captions at $d=5$ & $92.4$ \\
        \hline
    \end{tabular}
    \caption{The percentage of our creative captions at each degree of abstraction that our annotators judge as exhibiting visual grounding $\geq 3$ on our $4$-point Likert scale. Our captions demonstrate consistent alignment with their paired images, despite increasing abstraction.}
    \label{tab:caption_hallucination}
\end{table}

\begin{table}[h]
    \centering
    \begin{tabular}{|c|c|}
        \hline
        \textbf{Caption Type} & \textbf{Average Rank} \\ 
        \hline
        Original Captions & $1.47$ \\
        \hline
        Captions at $d = 1$ & $2.69$ \\
        \hline
        Captions at $d = 2$ & $3.39$ \\
        \hline
        Captions at $d = 3$ & $4.03$ \\
        \hline
        Captions at $d = 4$ & $4.50$ \\
        \hline
        Captions at $d = 5$ & $4.98$ \\
        \hline
    \end{tabular}
    \caption{The average abstraction rank (out of $6$) for MSCOCO's original and our creative captions. We find that as our specified degree of abstraction increases, annotators rank the resulting creative captions as exhibiting more abstraction, validating our method.}
    \label{tab:caption_abstraction}
\end{table}

\subsection{How contextualized are our associations?}

In Table \ref{tab:unique_associations}, we report the percentage of contextualized associations that are unique for each salient word (e.g. \textit{bear} in Figure \ref{fig:datasetexamples}) across all the images where it appears, averaged over all of the salient words in our corpus.  Across all levels of abstraction, each salient word's mined associations contain duplicates in less than $50$\% of cases. Moreover, as the degree of abstraction increases, these associations become increasingly unique. Thus, for a given image, the mined associations of its salient words are tailored to its specific visual context.

\subsection{How good are our creative captions?}

% In Table \ref{tab:caption_hallucination}, we show the results of our first human evaluation task, rating the visual grounding of our creative captions. 
In our first human evaluation, annotators rated the visual grounding of our creative captions on a four-point Likert scale.  Their judgments exhibit fair agreement (a Fleiss $\kappa$ of $0.303$) %for this visual grounding assessment 
as calculated from three-way annotation on 20\% of our tasks ($100$ captions, $5$ each across $20$ images) \cite{fleiss1971measuring}.  

We bucket the resulting labels into two groups, $(1, 2)$ indicating poor visual grounding, and $(3, 4)$ indicating acceptable visual grounding.  This bucketing threshold was chosen after analyzing captions that were assigned visual grounding scores of three (minor grounding errors) to characterize their issues.  We prompt a strong reasoning model, Gemini $2.0$ Flash, to describe errors in these captions that significantly alter the meaning of the image using the prompt in \ref{sec:appendix_error_analysis}.  We find that for all such captions, this VLM identified no actual errors.  Manual inspection suggests that annotators may have given lower scores to interpretive language -- i.e cases where the caption is neither incorrect nor literal to the image (see Figure \ref{fig:error_analysis_table}). Therefore, we consider any caption with a grounding score greater than or equal to three to be visually grounded.

% \new{To ensure the quality of our generated captions and verify our threshold for acceptable visual grounding, we conducted an analysis of captions where the majority of annotators gave a score of 3 indicating minor grounding errors. We prompted a high performing VLM with reasoning capabilities different from the models used in our paper, specifically Gemini 2.0 flash, to quantify the number of errors introduced in the caption that significantly alter the meaning of the image based on the prompt included in \ref{sec:appendix_error_analysis}. We find that for all examined captions, the error count identified by the VLM is 0. Upon further inspection of the explanations for number of errors detected shown in \ref{fig:error_analysis_table}, we believe our annotators might have given lower scores where interpretive language is used and the caption is neither incorrect nor literal to the image. Therefore, grounding scores that are reflected in our annotations are an underestimate of how grounded they are in the image.}

When considering our overall results in Table \ref{tab:caption_hallucination}, we can see that our creative captions demonstrate consistent visual alignment with their images -- in fact, at abstraction degrees $1$, $3$ and $5$, this is true of more than $90\%$ of our creative captions.  Our method for mining contextualized associations and generating creative captions generally avoids introducing errors and hallucinations.

% Table 2 shows the results of the second annotation task where annotators were asked to rate creative captions from distances of 1 to 5 for quality of the caption reflecting the given image. While annotators rated the captions on a scale of 1 to 4 we report the percentage at which ratings were greater than 2 to understand in a binary setting if the captions reflect the given the image. Overall, our results indicate that on average captions at all levels reflect their images with captions at distances 1,3,and 5 exceeding 0.9. We obtain inter-annotator agreement score of 0.303 (fair agreement) based on Fleiss's Kappa score. 

In Table \ref{tab:caption_abstraction}, we show the results of our second human evaluation task, ranking an image's captions (its original caption and a creative caption sampled for each of our five degrees of abstraction) in order of increasing abstraction.  We collect three annotations for 20\% of these tasks (totaling $120$ captions, $6$ captions each for $20$ images) and observe fair agreement (a Fleiss $\kappa$ of $0.283$) for this task given its subjective nature.  When considering our overall results, we can see the average rank of captions at each degree of abstraction reflects the intended abstraction, even relative to one another.  Captions at smaller degrees have lower rank than captions at higher degrees. Original captions receive the lowest average rank of $1.47$ and captions at $d=5$ receive the highest average rank of $4.98$. Our method is capable of consistently generating increasingly abstract creative captions.

% Overall, our results indicate that on average the rankings reflect the degree of their ground truth label distance labels. 

% We obtain inter-annotator agreement score of 0.283 (fair agreement) based on Fleiss's Kappa score\cite{fleiss1971measuring}.   

% \aanote{I am not sure we need this paragraph -- we can just say that we observe fair agreement for what are subjective tasks and perhaps compare to other published work -- I think Saif Mohammad has agreement for art emotion labels that is similar to what we're reporting here, for example.  If we go into too much detail hypothesizing why there's disagreement it undermines the results.}
% Although inter-annotater agreement scores 
% for both tasks are below moderate agreement, we attribute this to tasks of assessing quality of abstractness in images and texts being a rather subjective task especially between more abstract captions. For example in the first task it is easier to objectively choose which is more abstract between captions "A person on a surfboard about to hit a wave" (original caption) and caption "Patience allows a surfer to navigate life's waves with grace." (distance 5) But the task becomes more subjective when choosing between two abstract captions "Pausing at the wave's edge, a surfer rides time." (distance 4) and "Patience allows a surfer to navigate life's waves with grace." (distance 5). Similarly it is difficult to evaluate errors objectively in more abstract captions that do not directly describe the image. 

\begin{figure*}[h]
        \centering
        \includegraphics[width=\linewidth]{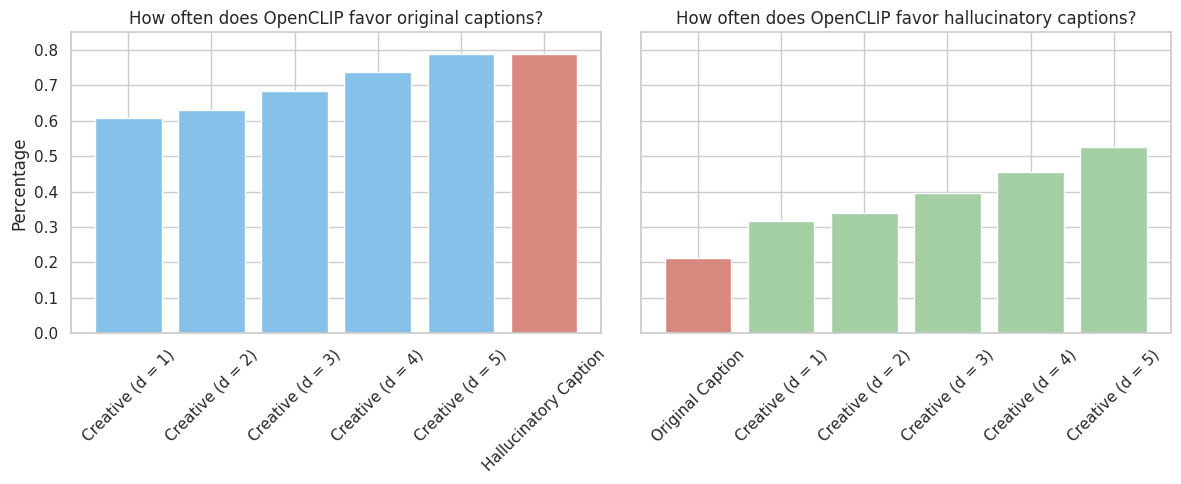}
        % \caption{Percentage of COCO original captions scoring higher than foil captions and creative captions in degrees 1 through 5 }
    \caption{Plots comparing OpenCLIP's scores for original (left) and hallucinatory (right) captions against its scores for our creative captions.  OpenCLIP favors literalism and cannot distinguish between hallucination and abstraction.}
    \label{fig:main}
\end{figure*}

\begin{table*}[h]
    \centering
    \begin{tabular}{lccccc}
        \hline
        
        \textbf{Model} & \textbf{$\mathbf{k=1}$ ($\uparrow$)} & \textbf{$\mathbf{k=5}$ ($\uparrow$)} & \textbf{$\mathbf{k=10}$ ($\uparrow$)} & \textbf{$\mathbf{k=20}$ ($\uparrow$)} & \textbf{Avg Rank ($\downarrow$)} \\
        \hline
        OpenCLIP       & $0.1505$ & $0.3089$ & $0.4033$ & $0.5043$ & $70.46$ \\
        OpenCLIP-FT ($d = 1$)     & $0.1454$ & $0.3086$ & $0.3934$ & $0.4977$ & $72.37$ \\
        OpenCLIP-FT ($d = 2$)     & $0.1446$ & $0.3048$ & $0.3913$ & $0.4877$ & $72.37$ \\
        OpenCLIP-FT ($d = 3$)     & $0.1485$ & $0.3106$ & $0.3935$ & $0.4912$ & $72.41$ \\
        OpenCLIP-FT ($d = 4$)     & $0.1591$ & $0.3233$ & $0.4150$ & $0.5147$ & $68.96^{*}$ \\
        OpenCLIP-FT ($d = 5$)    & $\mathbf{0.1624}$ & $\mathbf{0.3341}$ & $\mathbf{0.4162}$ & $\mathbf{0.5222}$ & $\mathbf{67.60}^{*}$\\
        \hline
    \end{tabular}
    \caption{\textbf{Task 1: Poetry-to-Image Retrieval.} 
 Recall@$k$ and average rank of OpenCLIP and a variant fine-tuned on our creative caption corpus at increasing degrees of abstraction.  At degrees $4$ and $5$, our fine-tuned model outperforms the baseline across all metrics.  * indicates significance at $\alpha=0.05$.}
    \label{tab:clip_recall}
\end{table*}

% \begin{table*}[h]
%     \centering
%     \begin{tabular}{lccccc}
%         \hline
%         \textbf{Model} & \textbf{Recall@1} & \textbf{Recall@5} & \textbf{Recall@10} & \textbf{Recall@20} & \textbf{Avg Rank} \\
%         \hline
%         CLIP Zero-Shot     & 0.00015  & 0.00107  & 0.00231  & 0.00324  & 3288.9  \\
%         Our CLIP P D = 1   & 0.000154 & 0.00062  & 0.00200  & 0.00416  & 3262.6  \\
%         Our CLIP P D = 2   & 0.00000  & 0.00062  & 0.001696 & 0.00324  & 3266.4  \\
%         Our CLIP P D = 3   & 0.00015  & 0.00077  & 0.00139  & 0.00339  & 3264.2  \\
%         Our CLIP P D = 4   & 0.000308 & 0.00108  & 0.00154  & 0.00278  & 3253.4  \\
%         Our CLIP P D = 5   & 0.000154 & 0.00092  & 0.00262  & 0.003547 & 3244.5  \\
%         \hline
%     \end{tabular}
%     \caption{Visual Metaphor Corpus Image to Text Retrieval zero shot CLIP vs CLIP with abstraction results \aanote{Perhaps for this we should just report the average rank numbers?  The extremely small recall numbers are very distracting.  In the main body we can say we don't report recall because it's very low / a challenging task}}
%     \label{tab:clip_recall_low}
% \end{table*}
\begin{table}[h!]
    \centering
    \begin{tabular}{lc}
        \hline
        \textbf{Model} & \textbf{Avg Rank ($\downarrow$)} \\
        \hline
        OpenCLIP &  $3288.9$  \\
        OpenCLIP-FT ($d = 1$)  & $3262.6$  \\
        OpenCLIP-FT ($d = 2$)  & $3266.4$  \\
        OpenCLIP-FT ($d = 3$)  & $3264.2$  \\
        OpenCLIP-FT ($d = 4$)  & $3253.4$  \\
        OpenCLIP-FT ($d = 5$)  & $\mathbf{3244.5}^{*}$  \\
        \hline
    \end{tabular}
    \caption{\textbf{Task 2: Visual Metaphor-to-Linguistic Metaphor Retrieval.}  The average rank of OpenCLIP and a variant fine-tuned on our creative caption corpus at increasing degrees of abstraction.  As abstraction increases, our model's average rank improves over the baseline.  * indicates significance at $\alpha=0.05$.}
    \label{tab:clip_recall_low}
\end{table}
\begin{table*}[h!]
    \centering
    \begin{tabular}{lc}
        \hline
        \textbf{Model Name} & \textbf{\% preference for DALL·E 2 (CoT)} $\uparrow$ \\
        \hline
        OpenCLIP     & 0.43 \\
        OpenCLIP-FT ($d = 1$)  & $\mathbf{0.59^{*}}$ \\
        OpenCLIP-FT ($d = 2$)   & $0.47$ \\
        OpenCLIP-FT ($d = 3$)   & $0.54^{*}$ \\
        OpenCLIP-FT ($d = 4$)   & $0.49$ \\
        OpenCLIP-FT ($d = 5$)  & $0.50$ \\
        \hline
    \end{tabular}
    \caption{\textbf{Task 3: Linguistic Metaphor-to-Visualization Matching.} Preference for the correct visualization of OpenCLIP and a variant fine-tuned on our creative caption corpus at increasing degrees of abstraction.  All abstraction settings improve over the baseline. * indicates significance at $\alpha=0.05$.}
    \label{tab:dalle_cot_comparison}
\end{table*}

\subsection{Does OpenCLIP agree?}

Our human evaluation makes clear that our synthetic creative captions are both visually grounded and exhibit recognizably increasing abstraction.  Does the original OpenCLIP model agree?  

To test this, for each image in our corpus, we calculate its similarity with its original caption and its similarity with its creative captions.  Additionally, as a baseline, we calculate its similarity with captions containing obvious hallucinations from the FOIL dataset \cite{shekhar2017foil}.  These hallucinatory captions specify concrete objects that are not present within each image, chosen from a present object's hypernym (e.g, replacing \textit{car} with \textit{bus}, another \textit{vehicle}).  Unlike our creative captions, these captions reflect poor visual grounding.

% \new{For this, a hallucinatory caption is chosen for each particular image based on the FOIL captions associated with each COCO image id. If multiple captions were present for a given image we randomly sampled from the given captions for the image. Hallucinatory captions in this case are captions from the FOIL dataset that contain objects that are not present within the image but belong to the same MSCOCO provided standard object categories of the original object being replaced.} %object categories are the class labels of the object (ex cafr belongs to vehicle that MSCOCO provides in their annotations  

On the left side of Figure \ref{fig:main}, we plot how often the original captions score higher than 1) our creative captions at increasing degrees of abstraction and 2) the FOIL captions. As the degree of abstraction increases, OpenCLIP favors the original captions more and more.  In fact, at our highest degree of abstraction, OpenCLIP prefers the original caption $80\%$  of the time, nearly the same rate at which it prefers the original caption over the hallucinatory captions from FOIL.  This suggests a strong preference for literal over creative captions.

On the right side of Figure \ref{fig:main}, we plot how often hallucinatory FOIL captions score higher than 1) original captions and 2) our creative captions at increasing degrees of abstraction.  As the degree of abstraction increases, it becomes more and more difficult for OpenCLIP to distinguish between obvious hallucinations and abstraction.  In fact, at our highest degree of abstraction, OpenCLIP does no better than random guessing. This shows that standard image-text datasets result in models unable to differentiate between hallucination and abstraction.   

% Together, these results demonstrate the difficulty that 

% Figure \ref{fig:main} shows the results of evaluating zero shot CLIP on our dataset. Figure \ref{fig:sub1} depicts that the percentage of times original captions score higher than creative captions from our dataset is greater than 50 percent in all degrees settings. As the degrees levels increase, indicating that the captions are increasingly becoming abstract, the original scores score increasingly higher than the creative captions. This is especially apparent with comparing the percentage of times original captions score higher than degree 5 captions and the percentage of times original captions score higher than foil captions (captions known to contain hallucinations) which are both at 0.8. This indicates zero shot CLIP has difficulty aligning abstract text and images compared to literal text and images  especially as texts become increasingly abstract. 

% Figure \ref{fig:sub2} depicts that the percentage of times foil captions score higher than creative captions from our dataset is consistently higher than the percentage of times foil captions score higher than the original captions. Similar to Figure 1a, we see that as the degrees increase and the captions become more abstract, the percentage of times foil captions score higher than the creative captions increases. Captions at degree 5 have the highest rate of scoring lower than foil captions. These results indicate zero shot CLIP has difficulty in differentiating between captions containing hallucinations and abstract captions when aligning with images. 
\subsection{Do contextualized associations improve downstream creative understanding?}

In Tables \ref{tab:clip_recall}, \ref{tab:clip_recall_low} and \ref{tab:dalle_cot_comparison}, we compare the performance of OpenCLIP against the performance of our model fine-tuned on our synthetic captions at all five degrees of abstraction across our selected creative understanding tasks.  In all three tasks, our creative captions yield significant improvements over the baseline despite no task-specific fine-tuning.

On poetry-to-image retrieval (Table \ref{tab:clip_recall}), our fine-tuned variant improves over the baseline in both recall and average rank when the degree of abstraction is set to either $4$ or $5$, with $5$, our highest degree of abstraction, exhibiting the best performance.

On visual metaphor-to-textual metaphor retrieval, both zero-shot CLIP and our fine-tuned variant struggle to achieve reasonable recall values.  However, when we plot the average rank of the correct textual metaphor, we see that increasing the degree of abstraction in our fine-tuned visual encoder yields consistent reductions.

% increasing the degree of abstraction yields consistent reductions in average rank

% we 

% Table \ref{tab:clip_recall} shows the results of our CLIP model fine-tuned on our dataset with prefix embeddings tuning only, on the cross modal task of poem to image retrieval. Our results show that overall our CLIP model in more abstract degree settings of 4 and 5, achieves higher recall values at all k and lower average rank for the correct poem image pair than zero-shot CLIP.

% Table \ref{tab:clip_recall_low}  shows the results of our CLIP model fine-tuned on our dataset with prefix embeddings tuning only, on the cross modal task of visual metaphor image to text retrieval. Our results show that overall both zero-shot CLIP and our CLIP models struggle to achieve high recall at k scores for all k values. While all our models do achieve slightly lower average rank scores for all degree levels, the best performing models are at more abstract degrees of 4 and 5.    

On linguistic metaphor-to-visualization matching (Table \ref{tab:dalle_cot_comparison}), our fine-tuned variant improves over the baseline at every degree of abstraction.  Interestingly, we observe the largest improvement at a relatively low degree of abstraction ($d = 1$), a break with prior trends.  

% Table \ref{tab:dalle_cot_comparison} shows the results of our CLIP model fine-tuned on our dataset with prefix embeddings tuning only, on the model preference task of choosing the better Visual Metaphor given a linguistic metaphor. Our results show that overall our models outperform zero-shot CLIP in all settings with best performance at degree 1.    

% From our results, we see that for two out of three tasks, our fine-tuned models do exceed performance of zero-shot CLIP. However, for the task of Visual Metaphor Retrieval, the recall @k values for all models are low which suggests the task is hard for all models including our fine-tuned models. 

In making sense of the differences among our model's performances across all three tasks, we hypothesize that one important source of variation could be the composition of our evaluation data. Much like our synthetic corpus, which contains creative captions paired with ordinary images, MultiM-Poem contains figurative language paired with photographs from Flickr.  This poses a smaller domain shift than HAIVMet, where creative language is paired with creative imagery. 

We also note performance variation across abstraction levels. One reason for this is we do not know the right level of abstraction for a particular creative task a priori.  This highlights the need for corpus generation techniques like ours where the desired degree of abstraction can be specified. 
%We observed that some tasks such as Visual Metaphor-to-Linguistic Metaphor Retrieval and Linguistic Metaphor-to-Visualization Matching require more abstraction on the visual side as opposed to the text side. Meanwhile the Poetry-to-Image Retrieval tasks rely more heavily on the text side for abstraction. 
% These task-specific differences are reflected in our results, highlighting the need for corpus generation techniques like ours where the desired level of abstraction can be specified.

Nevertheless, given the improvements exhibited by our fine-tuned model  and the relative ease of applying our corpus generation technique to other image corpora, we view our results as strong evidence for the value of our mined associations in adapting vision-language understanding to creative domains.

% One explanation for this is that while our corpus does contain elements of figurative language and our prefix embedding allows for the models to associate text at increasingly abstract levels, the visual metaphor retrieval task is conducted on images and text pairs in which both the image and text are abstract. Our corpus is generated from images from the MSCOCO corpus which contain realistic images that we pair with abstract text. This is similar to the poetry image corpus which contains realistic images with abstract text and thus our model outperforms zeroshot clip in increasingly abstract settings. Meanwhile the HAIVMET corpus contains synthetic artistic images generated from explanations of linguistic metaphors which is inherently different from our image corpus. This could have contributed to low performance across the recall metrics in all models. T
% hus captions containing conceptual associations at increasingly abstract levels do increase performance on downstream creative image text pairs, but these gains are limited depending on how divergent the images are from our corpus. One possible way to remedy this is to scale our dataset on synthetically generated images from our captions and see if performance improves.  

\section{Conclusion \& Future Work}

In this work, we introduce a scalable method for mining contextualized associations for visual elements that can be applied to any corpus of unlabeled images.  We use these associations to produce a new dataset of increasingly abstract creative captions for MSCOCO.  Both human judgment and automatic evaluation across three challenging image-language tasks confirm the value of this method for enabling creativity understanding.  In the future, we plan to extend this study beyond English and Western associations -- recent work has shown that, in some cases, VLMs exhibit culturally specific regularities when prompted in different languages \cite{ananthram2024see}.  It is our hope to leverage this to mine multicultural associations at scale.

% We conduct human evaluations to determine the quality of our creative captions and find our dataset reflects the increasing abstractness associated with their degree labels and are recognizably descriptive of their paired images. We show using our dataset to fine-tune CLIP leads to improvement in downstream cross modal retrieval tasks in poetry image pairs and Visual Metaphors corpora in zero shot settings. Future work could include exploring augmenting other datasets with contextualized associations as well as evaluating in generation tasks to see the effect of associations on text or visual generations.   

\section{Limitations}

While our method for mining visual associations and generating creative captions is easy to scale, we acknowledge its reliance on \texttt{gpt4o-mini}, a paid closed source model. Additionally, we use \texttt{Molmo} to generate both the detailed descriptions and the creative descriptions of the images in our corpus.  LLMs and VLMs are both prone to hallucinations and biases which could be reflected and reinforced by both our method and our dataset. 
%While we evaluate CLIP trained on our corpora against zero-shot CLIP, we do not evaluate other models or CLIP fine-tuned on creative captions directly prompted using gpt4o-mini or another LLM.
Moreover, there is room for improvement across all evaluation tasks which can be achieved through using additional datasets including more variation in images and captions as well as other prompting techniques that have not been explored in this work. Finally, our contextualized associations are limited to the English language and likely reflect a Western-centric perspective. However our methods allows for scalability in other languages which can be conducted in future work.  

\section{Ethics Statement}
Our work relies on large-language models and vision-language models which are known to have potential biases and limited interpretability that can be harmful. However, we focus on using our resulting datasets and methods as research tools to support the study of creativity and abstraction in vision-language models. Thus we have no additional concerns regarding harmful use other than those already present in the broader field of generative artificial intelligence. 

Human evaluations were conducted to assess the quality of the generated captions. All participants were recruited voluntarily and provided information about the intended use of their annotations for research purposes. No personally identifiable information was collected and all responses were stored and analyzed in anonymized form. Demographic and geographic information on annotators was not collected or reported to ensure full anonymity.

\bibliography{custom}
\newpage
\appendix

\section{Appendix}
\renewcommand{\thefigure}{A.\arabic{figure}}
\setcounter{figure}{0} % reset figure counter
\label{sec:appendix}

\subsection{Computation and Model Specifics}
The base CLIP model has around 86 million parameters. Our CLIP model variant was trained on 1 NVIDIA RTX A6000 GPU for 1 epoch taking roughly 3 hours. We used a learning rate of \texttt{1e-4} and \texttt{torch.optim.Adam} optimizer. We use early stopping and lowest validation loss with a \texttt{patience = 3} on our synthetic corpus validation dataset to determine the best model. \\

The specific version of Spacy we use is spacy $3.8.4$\\

We use the OpenAI batch API to generate our associations. The specific hyperparemters we use apart from defaults is \texttt{max\_tokens: 1000} \\

The hyperparameters we use for Molmo is \texttt{temperature=0.7}, \texttt{top\_p=0.9}, \texttt{max\_tokens=150}, \texttt{n=1}. 
\subsection{Generating Abstracted Captions}
\label{sec:appendix_mining}

\subsubsection{Detailed Caption Prompt}
Below is the prompt used for generating the detailed caption of a given image.\\ 

""USER: <image> Please generate a detailed caption of this image. ASSISTANT:"
\label{sec:appendix_detailed_caption}

\subsubsection{Mining Associations Prompt}
We prompt \texttt{GPT-4o-mini} using the batch API with the following system prompt where \{context\_caption\} refers to the detailed caption generated for an image and \{original\_caption\} refers to the MSCOCO caption of the image:\\

% \begin{listing}
% "For a given list of words, generate a new list for each word
% using the same part of speech. The words should follow a
% semantic abstraction scale where degrees increases from
% near-synonyms to abstract concepts.\\
% Approach:\\
% 1. Degree 1 – Near Synonyms: Close in meaning or form 
%    (e.g., Ball $\rightarrow$ Sphere).
% 2. Degree 2 – Slight Abstraction: Slightly broader category 
%    (e.g., Ball $\rightarrow$ Toy).
% 3. Degree 3 – Broader Context: Indirectly linked through 
%    situational and emotional context (e.g., Ball $\rightarrow$ Game).
% 4. Degree 4 – Conceptual Association: More abstract or 
%    theme-related (e.g., Ball $\rightarrow$ Competition).
% 5. Degree 5 – Full Abstraction: Highly abstract or 
%    metaphorical (e.g., Ball $\rightarrow$ Journey).

% Generate three words each for degrees 1 to 5. Generated 
% words should fit into the overall emotional and situational 
% context of this context caption: \{context\_caption\}

% Generated words, when replaced with the original word in 
% this short caption \{original\_caption\}, should be semantically 
% correct.

% Do not generate the original word in the new generations. 
% Use JSON format: the key is the original word, and the value 
% is a dictionary with degrees as keys and lists of generated 
% words as values."
% \end{listing}
\noindent
"For a given list of words, generate a new list for each word
using the same part of speech. The words should follow a
semantic abstraction scale where degrees increase from
near-synonyms to abstract concepts.

\textbf{Approach:}

\begin{enumerate}
  \item \textbf{Degree 1 – Near Synonyms:} Close in meaning or form 
  (e.g., Ball $\rightarrow$ Sphere).
  
  \item \textbf{Degree 2 – Slight Abstraction:} Slightly broader category 
  (e.g., Ball $\rightarrow$ Toy).
  
  \item \textbf{Degree 3 – Broader Context:} Indirectly linked through 
  situational and emotional context (e.g., Ball $\rightarrow$ Game).
  
  \item \textbf{Degree 4 – Conceptual Association:} More abstract or 
  theme-related (e.g., Ball $\rightarrow$ Competition).
  
  \item \textbf{Degree 5 – Full Abstraction:} Highly abstract or 
  metaphorical (e.g., Ball $\rightarrow$ Journey).
\end{enumerate}

Generate three words each for degrees 1 to 5. Generated 
words should fit into the overall emotional and situational 
context of this context caption: \{context\_caption\}.

Generated words, when replaced with the original word in 
this short caption \{original\_caption\}, should be semantically 
correct.

Do not generate the original word in the new generations.

\textbf{Output format:} Use JSON. The key is the original word, 
and the value is a dictionary with degrees as keys and lists 
of generated words as values."

\label{sec:appendix_associations_prompt}

\subsubsection{Abstracted Caption Prompt}
Below is the prompt used to obtain creative captions for an image for each of its salient objects and associations generated. \{all\_words\} contain the salient words for the image with the original word replaced with the association word at degree \{level\}. \{new\_word\} is the association word at degree \{level\}.  <image> is the input image\\

% \begin{listing}
% "USER: <image>
% Write a short caption grounded in this image and 
% semantically correct, using fewer than 10 words. 
% Choose some or all of these words: \{all\_words\} 
% to best represent the image. 

% Steer the caption's style toward the abstraction 
% level \_{label\_} following these rules:

% - Degree 1: Near Synonyms – Close in meaning 
%   to the original image
% - Degree 2: Slight Abstraction – Slightly more 
%   abstract than the image
% - Degree 3: Broader Context – Indirectly linked 
%   through situational or emotional context
% - Degree 4: Conceptual Association – More abstract, 
%   theme-related to the image
% - Degree 5: Full Abstraction – Highly abstract 
%   or metaphorical

% The caption MUST include the word: \{new\_word\}.
% ASSISTANT:"
% \end{listing}
\noindent
"USER: \texttt{<image>} \\
Write a short caption grounded in this image and 
semantically correct, using fewer than 10 words. 
Choose some or all of these words: \{all\_words\} 
to best represent the image.

Steer the caption's style toward the abstraction 
level \_{label\_} following these rules:

\begin{itemize}
  \item \textbf{Degree 1 – Near Synonyms:} Close in meaning to the original image
  \item \textbf{Degree 2 – Slight Abstraction:} Slightly more abstract than the image
  \item \textbf{Degree 3 – Broader Context:} Indirectly linked through situational or emotional context
  \item \textbf{Degree 4 – Conceptual Association:} More abstract, theme-related to the image
  \item \textbf{Degree 5 – Full Abstraction:} Highly abstract or metaphorical
\end{itemize}

The caption \textbf{MUST} include the word: \{new\_word\}.

\noindent
ASSISTANT:"

\label{sec:appendix_abstracted_caption}

\subsection{Evaluation}

\subsubsection{Significance Tests}
We use pairwise t-tests to report significance results on the results of task 3 involving a pairwise preference of images. Specifically we use \texttt{scipy.stats ttest\_rel} implementation\\ 

We use wilcoxin tests to report significance results on task 1 and 2 involving average ranks of the correctly retrieved image/text. Specifically we use \texttt{scipy.stats wilcoxon} implementation 
\subsubsection{Annotators and Annotation Interfaces}
We do not report demographics of annotators to maintain full anonymity. Collected annotator data are fully anonymized. Annotators were informed of their annotations would be used for research purposes. Below are the instructions and interfaces annotators used to complete the annotations tasks.    
\label{sec:appendix_interface}

\begin{figure*}[!ht]
    \centering
    \includegraphics[width=2\columnwidth]{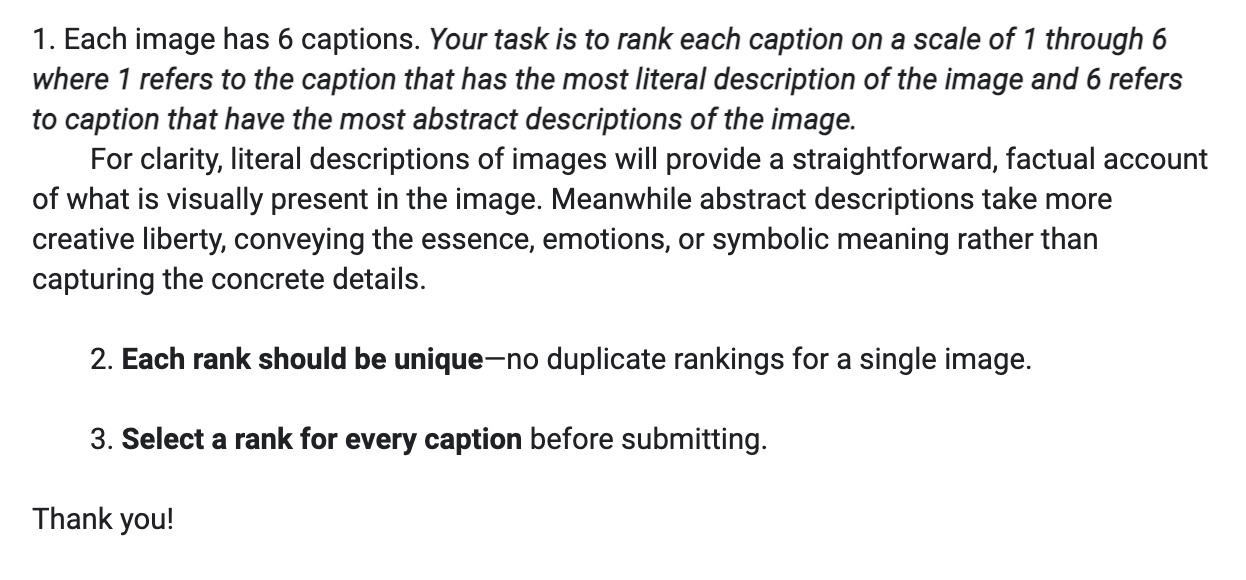}
    \caption{Instruction given to annotators for task 1 of Human Evaluation} 
    \label{fig:instructions1}
\end{figure*}

\begin{figure*}[!ht]
    \centering
    \includegraphics[width=2\columnwidth]{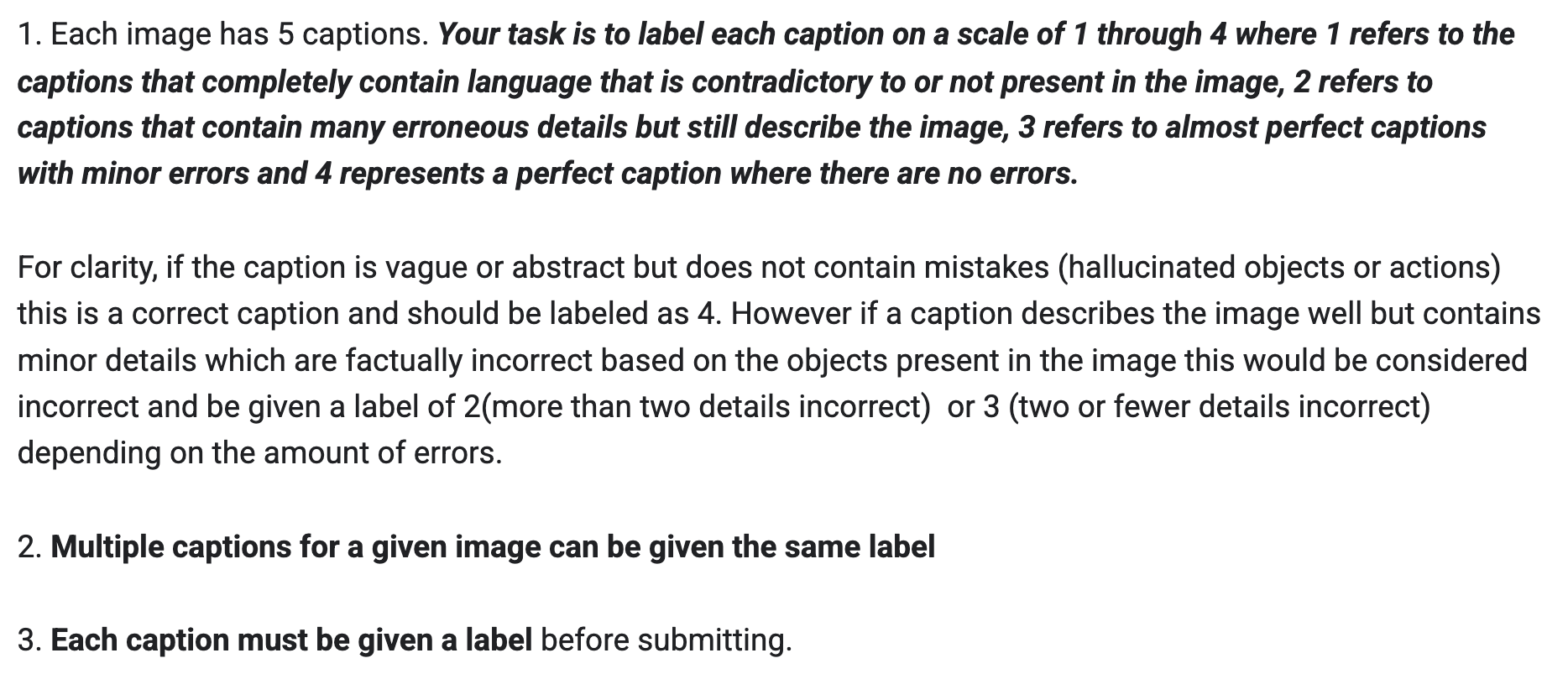}
    \caption{Instruction given to annotators for task 2 of Human Evaluation} 
    \label{fig:instructions2}
\end{figure*}
\begin{figure*}[!ht]
    \centering
    \includegraphics[width=1.5\columnwidth]{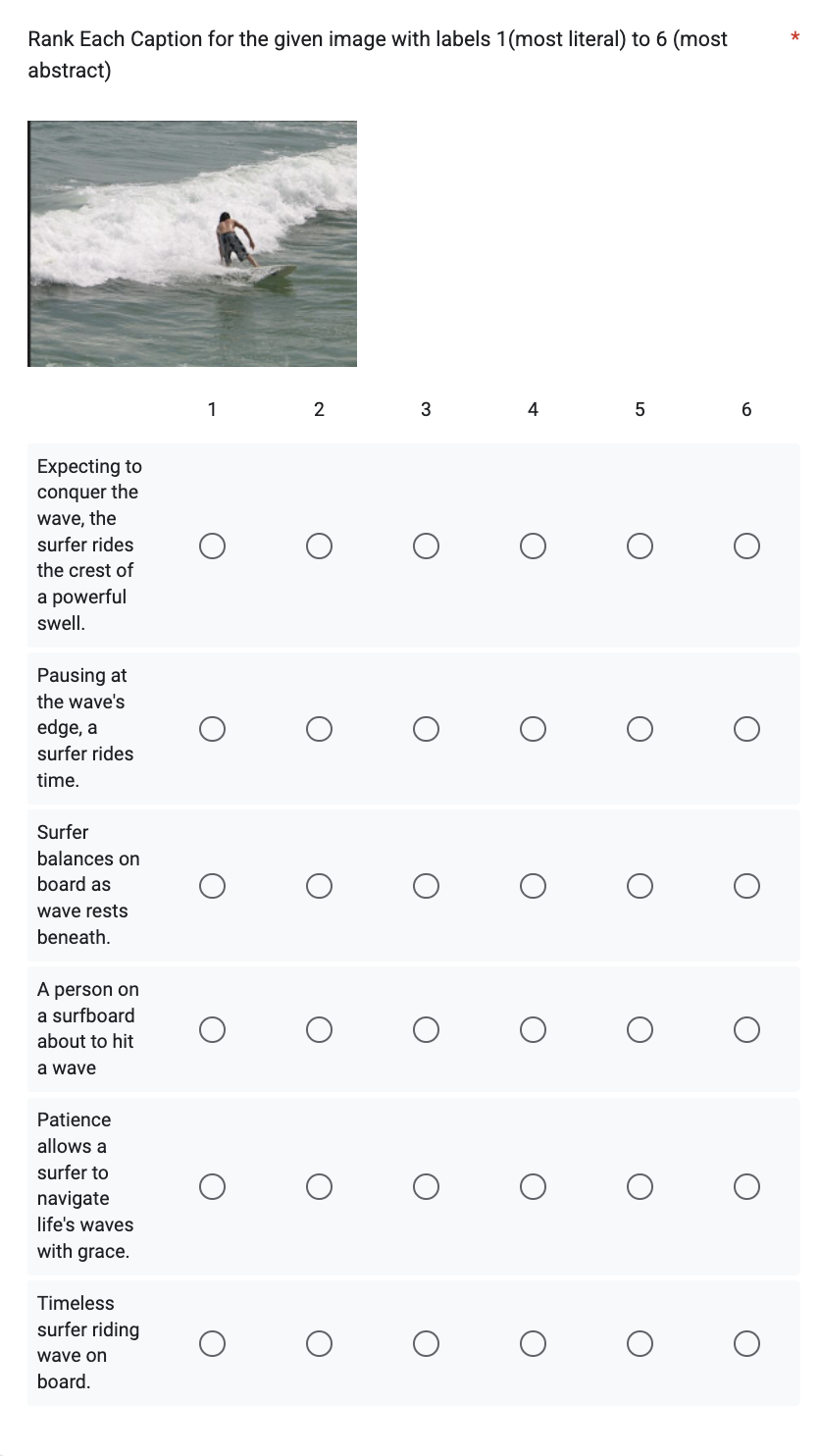}
    \caption{Example interface for one annotation from task 1} 
    \label{fig:annotation1}
\end{figure*}
\begin{figure*}[ht]
    \centering
    \includegraphics[width=1.5\columnwidth]{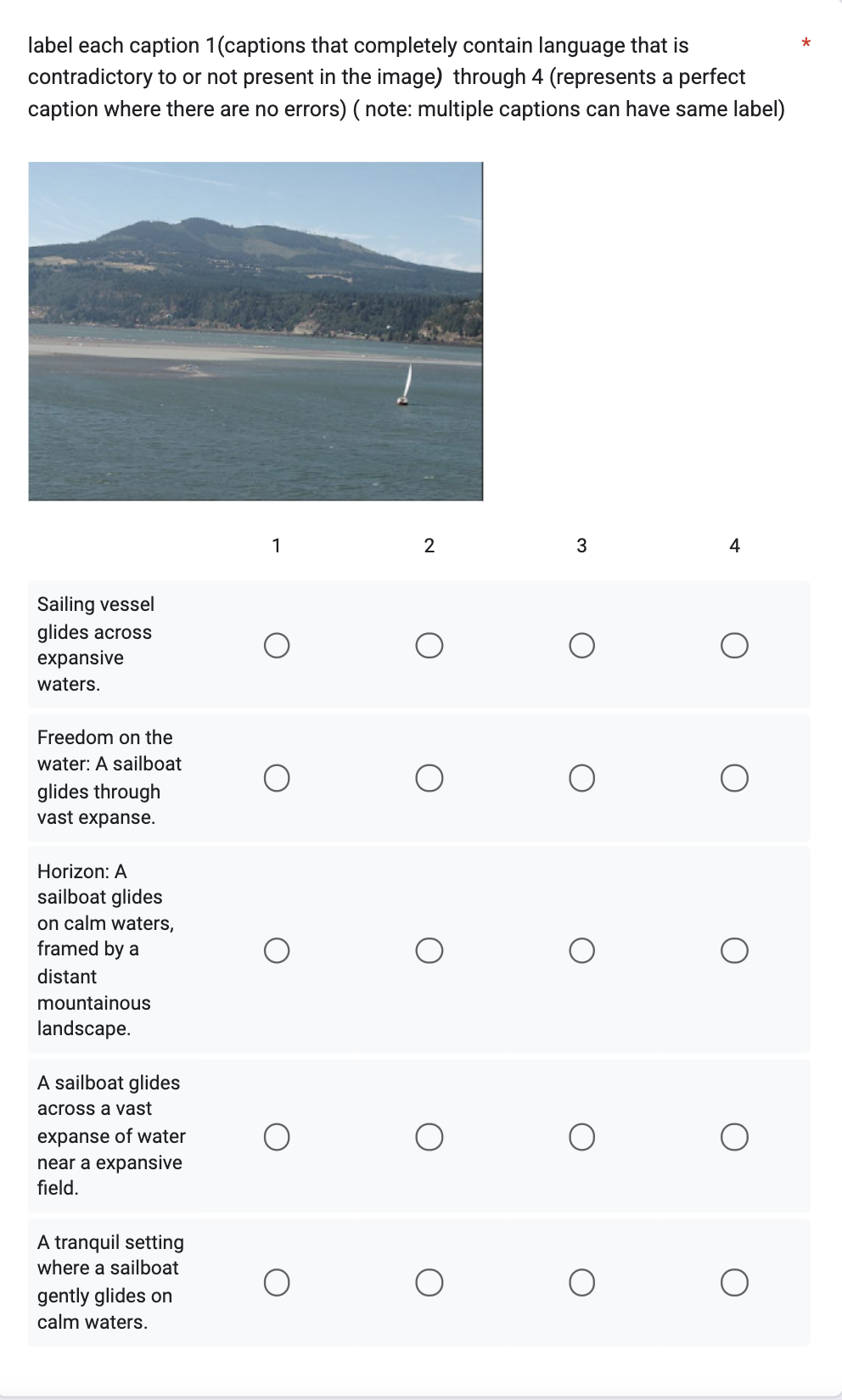}
    \caption{Example interface for one annotation from task 2} 
    \label{fig:annotations2}
\end{figure*}

% \subsection{Example Abstract Captions}
\label{sec:appendix_example_captions}
\begin{figure*}[htbp]
\centering
\renewcommand{\arraystretch}{1.5}
\begin{tabular}{p{0.25\textwidth}|p{0.35\textwidth}|p{0.35\textwidth}}
\textbf{Image} & \textbf{Original Caption} & \textbf{Creative Caption (D=5)} \\
\hline

% ===== Row 1 =====
\begin{minipage}{0.25\textwidth}\centering
    \includegraphics[width=\textwidth]{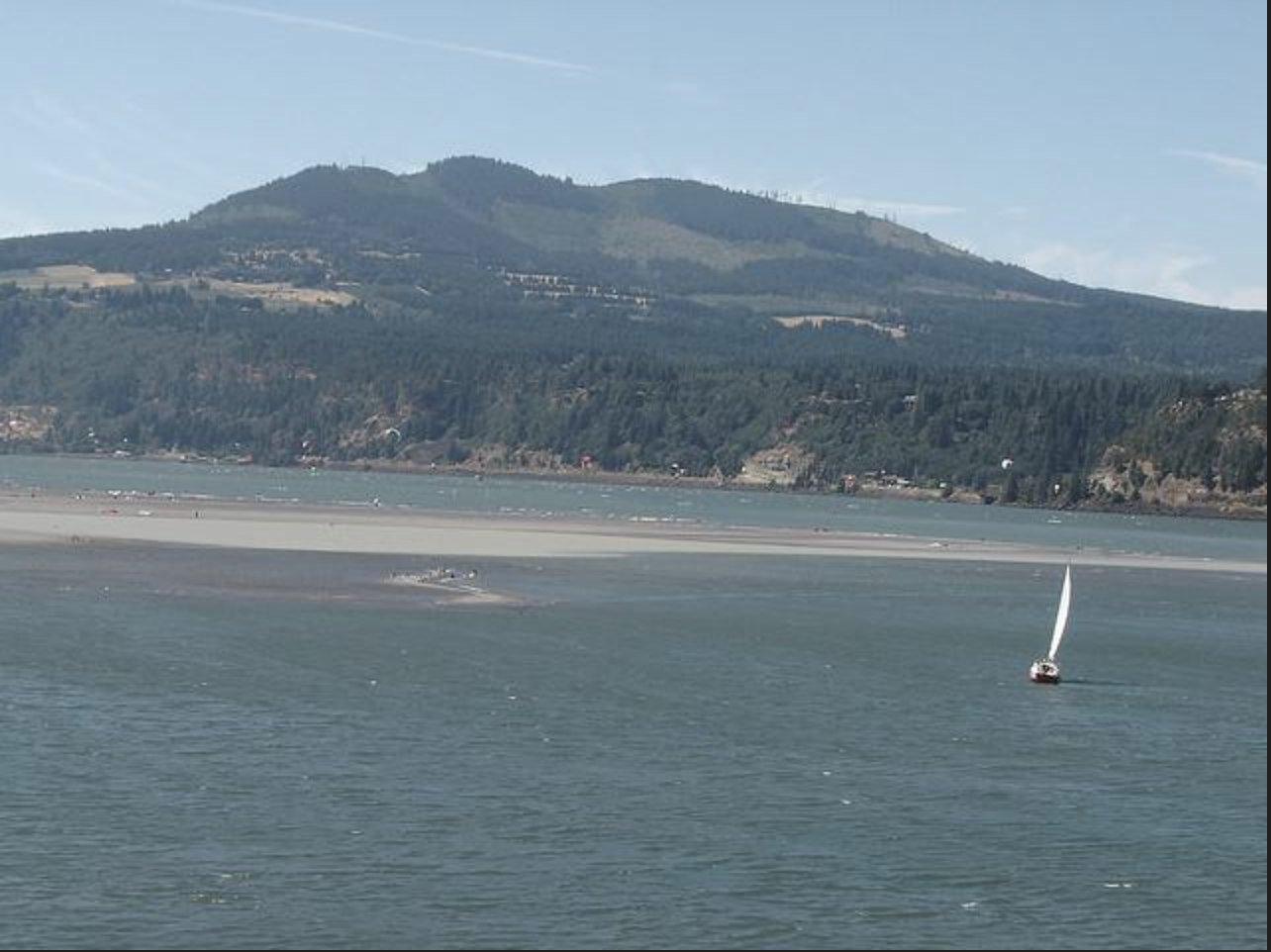}
\end{minipage} &
\begin{minipage}{0.35\textwidth}
    \textit{A small white sailboat in open water heading toward shore.}
\end{minipage} &
\begin{minipage}{0.35\textwidth}
    \textit{Freedom on the water: A sailboat glides through vast expanse.}
\end{minipage} \\
\hline

% ===== Row 2 =====
\begin{minipage}{0.25\textwidth}\centering
    \includegraphics[width=\textwidth]{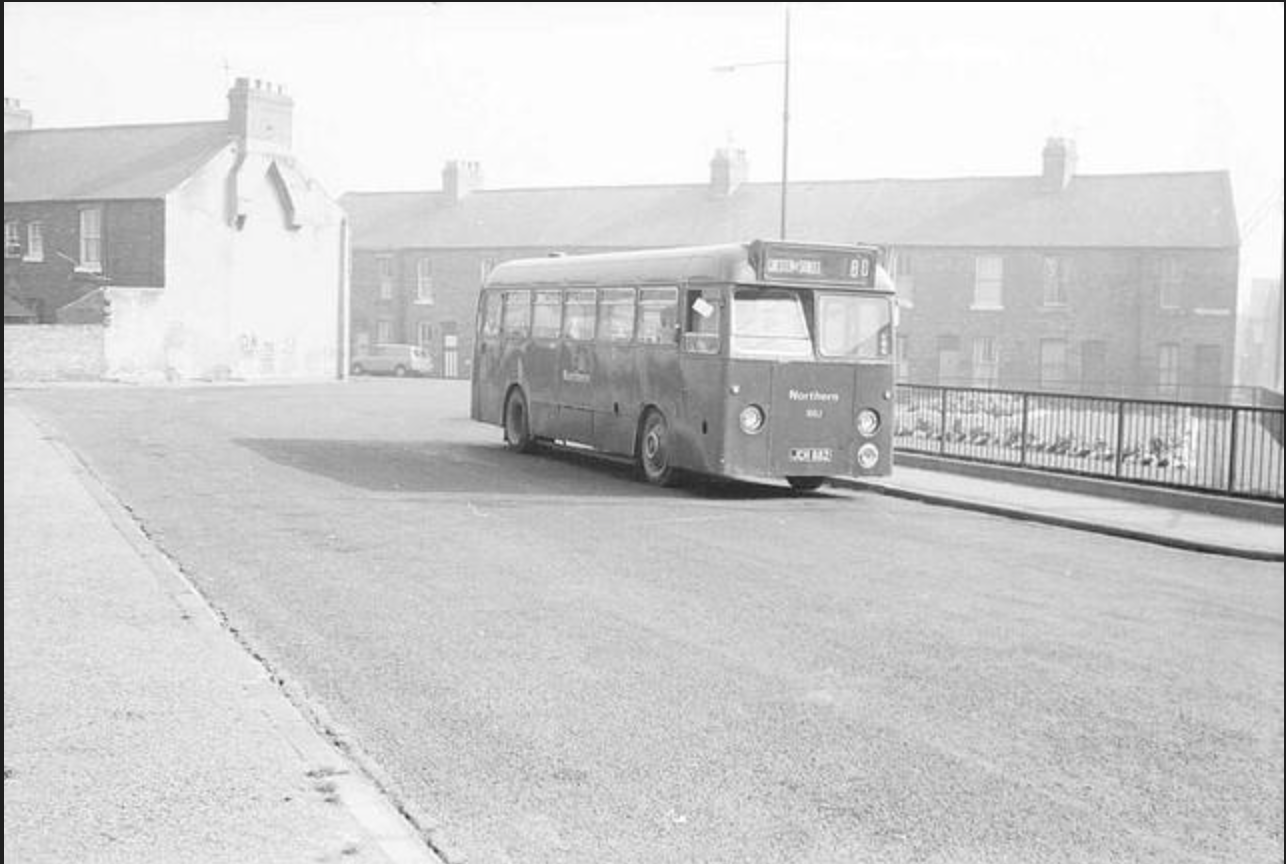}
\end{minipage} &
\begin{minipage}{0.35\textwidth}
    \textit{An old fashioned bus parked by the side of the road.}
\end{minipage} &
\begin{minipage}{0.35\textwidth}
    \textit{A journey begins in this town's serene landscape.}
\end{minipage} \\
\hline

% ===== Row 3 =====
\begin{minipage}{0.25\textwidth}\centering
    \includegraphics[width=\textwidth]{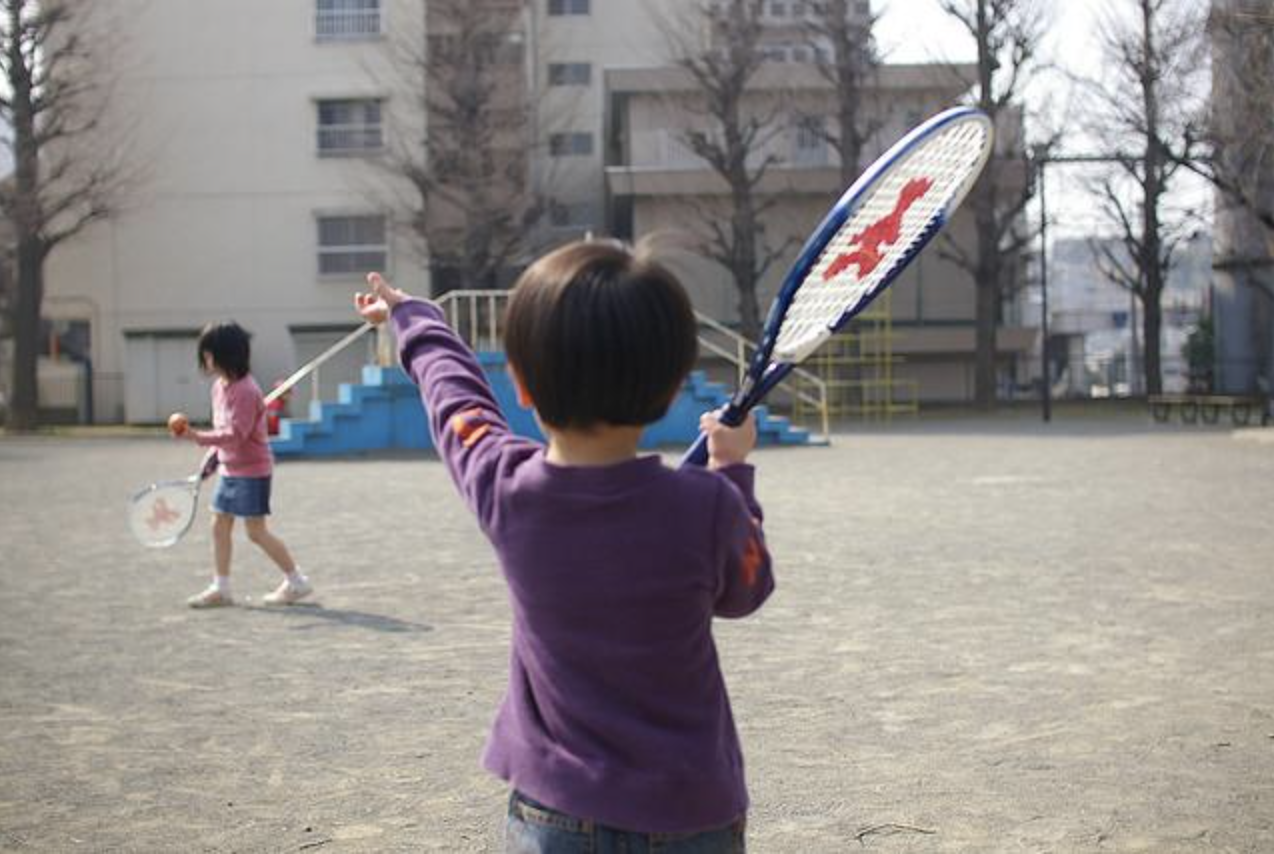}
\end{minipage} &
\begin{minipage}{0.35\textwidth}
    \textit{A boy and girl are playing tennis in the park.}
\end{minipage} &
\begin{minipage}{0.35\textwidth}
    \textit{Young minds chasing dreams: Children's determination on a playground tennis court.}
\end{minipage} \\
\hline

% ===== Row 4 =====
\begin{minipage}{0.25\textwidth}\centering
    \includegraphics[width=\textwidth]{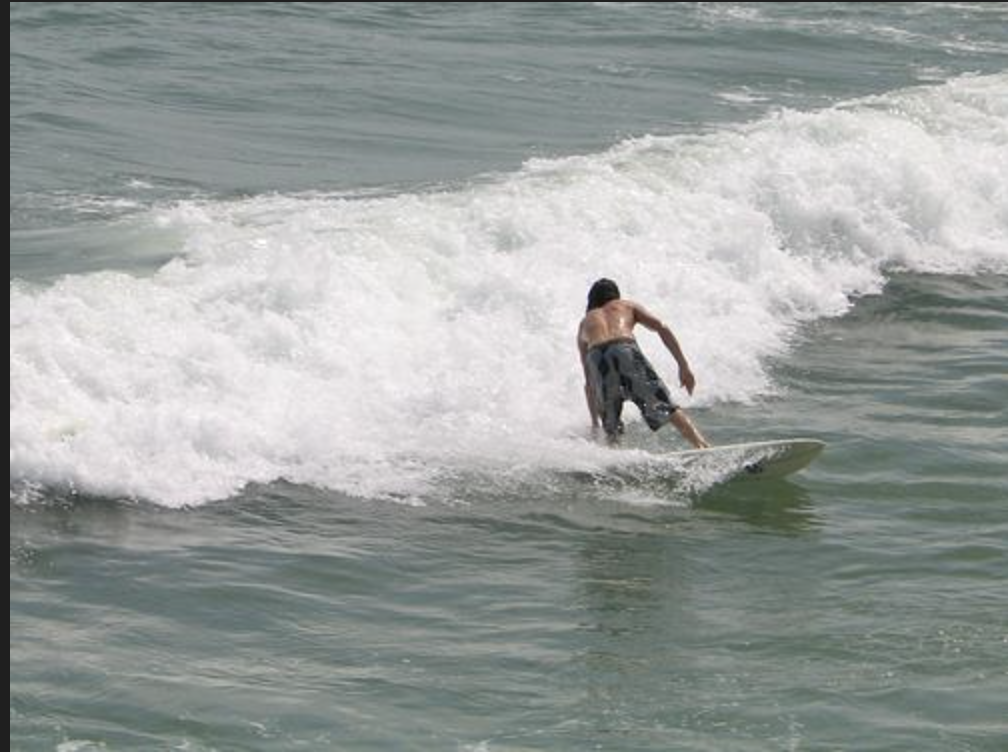}
\end{minipage} &
\begin{minipage}{0.35\textwidth}
    \textit{A person on a surfboard about to hit a wave.}
\end{minipage} &
\begin{minipage}{0.35\textwidth}
    \textit{Pausing at the wave's edge, a surfer rides time.}
\end{minipage} \\

\end{tabular}
\caption{Qualitative examples of captions for COCO images at most abstract degree level (d = 5)}
\label{fig:abstract_caption_examples}
\end{figure*}

\subsection{Error Analysis}
\label{sec:appendix_error_analysis}
Error analysis prompt: “[object Object] For this image count the number of errors for each sentence where errors are mistakes that significantly alter meaning of the image. Abstract elements are not errors if they do not alter the meaning of the image. Give an explanation for each sentence's score.” 
\begin{figure*}[htbp]
\centering
\renewcommand{\arraystretch}{1.5}
% \begin{tabular}{c|c|c}
\begin{tabular}{p{0.25\textwidth}|p{0.35\textwidth}|p{0.35\textwidth}}
\textbf{Image} & \textbf{Captions} & \textbf{Explanations} \\
\hline

% ===== Row 1 =====
\begin{minipage}{0.25\textwidth}\centering
    \includegraphics[width=\textwidth]{image1.png}
\end{minipage} &
\begin{minipage}{0.35\textwidth}
    \textbf{COCO:} \textit{people with umbrellas in a city at night.} \\
    \textbf{Creative:} \textit{poised figures with umbrellas navigating a rainy night in the city.} \\
    \textbf{Errors:} 0
\end{minipage} &
\begin{minipage}{0.35\textwidth}
\begin{itemize}[leftmargin=*, topsep=5pt, itemsep=3pt]
  \item ``poised figures'': Subjective interpretation implying elegance, not necessarily primary impression.
  \item ``umbrellas'': Accurate; people clearly holding umbrellas.
  \item ``navigating'': Accurate; figures are moving through the scene.
  \item ``rainy night'': Accurate; wet street and dark setting imply rain at night.
  \item ``in the city'': Accurate; urban buildings visible.
\end{itemize}
\end{minipage} \\
\hline

% ===== Row 2 =====
\begin{minipage}{0.25\textwidth}\centering
    \includegraphics[width=\textwidth]{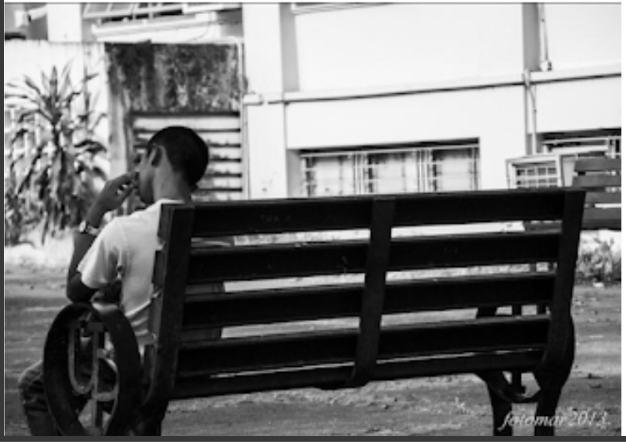}
\end{minipage} &
\begin{minipage}{0.35\textwidth}
    \textbf{COCO:} \textit{Man in T-shirt sitting on one end of a wooden bench, in an urban area.} \\
    \textbf{Creative:} \textit{man's posture suggests contemplation on wooden urban bench.} \\
    \textbf{Errors:} 0
\end{minipage} &
\begin{minipage}{0.35\textwidth}
\begin{itemize}[leftmargin=*, topsep=5pt, itemsep=3pt]
  \item ``man's posture'': Accurate; figure positioned reflectively.
  \item ``suggests contemplation'': Accurate interpretation of posture and composition.
  \item ``wooden urban bench'': Accurate; bench is wooden and urban.
\end{itemize}
\end{minipage} \\
\hline

% ===== Row 3 =====
\begin{minipage}{0.25\textwidth}\centering
    \includegraphics[width=\textwidth]{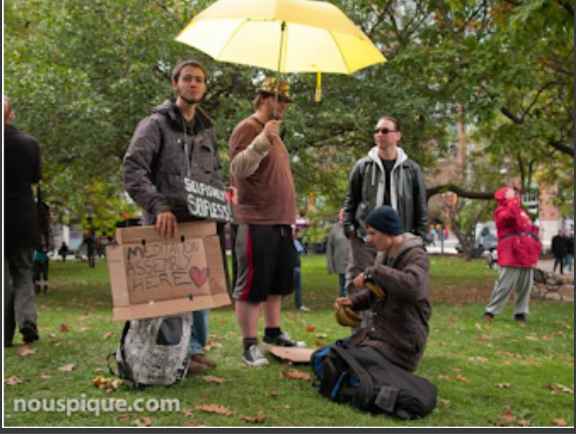}
\end{minipage} &
\begin{minipage}{0.35\textwidth}
    \textbf{COCO:} \textit{A group of men holding a Meditation sign and an umbrella.} \\
    \textbf{Creative:} \textit{guys practice patience during outdoor meditation assembly in park setting.} \\
    \textbf{Errors:} 0
\end{minipage} &
\begin{minipage}{0.35\textwidth}
\begin{itemize}[leftmargin=*, topsep=5pt, itemsep=3pt]
  \item ``guys'': Accurate; image shows a group of men.
  \item ``practice patience'': Subjective interpretation based on context.
  \item ``outdoor meditation assembly'': Accurate; sign and setting confirm this.
  \item ``park setting'': Accurate; background shows trees and grass.
\end{itemize}
\end{minipage} \\

\end{tabular}
\caption{ImageGrounding Error analysis examples. Each row shows the image, the COCO and Creative captions with annotated number of errors, and explanations for each element from Gemini 2.0 Flash model}
\label{fig:error_analysis_table}
\label{fig:error_analysis_table}
\end{figure*}

\end{document}